\documentclass[11pt]{article}

\usepackage[preprint]{acl}

\usepackage{times}
\usepackage{latexsym}

\usepackage[T1]{fontenc}

\usepackage[utf8]{inputenc}

\usepackage{microtype}

\usepackage{inconsolata}

\usepackage{graphicx}
\usepackage{amssymb}            
\usepackage{mathtools}          
\usepackage{mathrsfs}           
\usepackage{graphicx}           
\usepackage{subcaption}         
\usepackage[space]{grffile}     
\usepackage{url}                
\usepackage{lipsum}             

\usepackage[utf8]{inputenc}
\usepackage[T1]{fontenc}
\usepackage{textcomp}

\usepackage{amsmath, amsfonts, amssymb}
\usepackage{amsthm} 

\usepackage{booktabs}
\usepackage{graphicx}

\usepackage{algorithm}
\usepackage{algpseudocode}

\newtheorem{theorem}{Theorem}
\newtheorem{lemma}{Lemma}
\newtheorem{proposition}{Proposition}

\newtheorem{assumption}{Assumption} 

\usepackage{bbm}

\usepackage{booktabs, colortbl, xcolor, multirow, array, makecell, rotating, graphicx, amsmath}
\definecolor{bestcell}{RGB}{198,239,206}      
\definecolor{secondcell}{RGB}{255,235,156}    
\definecolor{headbg}{RGB}{68,114,196}         
\definecolor{headfg}{RGB}{255,255,255}        
\definecolor{rowalt}{RGB}{242,242,242}        
\definecolor{ourrow}{RGB}{220,230,242}        
\definecolor{warnred}{RGB}{255,199,206}       

\definecolor{bestblue}{RGB}{26,111,175}
\definecolor{secondamber}{RGB}{180,95,6}
\definecolor{worstred}{RGB}{180,30,30}
\definecolor{notegrey}{RGB}{120,120,120}
\definecolor{gaingreen}{RGB}{34,139,34}

\newcommand{\best}[1]{\textcolor{bestblue}{\textbf{#1}}}
\newcommand{\second}[1]{\textcolor{secondamber}{\textbf{#1}}}
\newcommand{\worst}[1]{\textcolor{worstred}{\textit{#1}}}
\newcommand{\noteg}[1]{\textcolor{notegrey}{\textit{#1}}}
\newcommand{\gain}[1]{\textcolor{gaingreen}{\textbf{#1}}}
\newcolumntype{C}[1]{>{\centering\arraybackslash}p{#1}}
\newcolumntype{L}[1]{>{\raggedright\arraybackslash}p{#1}}

\title{CoGate-LSTM: Prototype-Guided Feature-Space Gating for Mitigating Gradient Dilution in Imbalanced Toxic Comment Classification}

\author{
\begin{tabular}{c}
\small Noor Islam S.\ Mohammad$^{1,*}$ \\
\tt\small islam23@itu.edu.tr \\
\includegraphics[height=0.8cm]{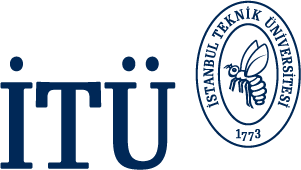} \\
\small $^{1}$Istanbul Technical University \\
\end{tabular}
}

\begin{document}
\maketitle

\begin{abstract}
Toxic text classification for online moderation remains challenging under extreme class imbalance, where rare but high-risk labels such as \emph{threat} and \emph{severe\_toxic} are consistently underdetected by conventional models. We propose \textbf{CoGate-LSTM}, a parameter-efficient recurrent architecture built around a novel \emph{cosine-similarity feature gating} mechanism that adaptively rescales token embeddings by their directional similarity to a learned toxicity prototype. Unlike token-position attention, the gate emphasizes feature directions most informative for minority toxic classes. The model combines frozen multi-source embeddings (GloVe, FastText, and BERT-CLS), a character-level BiLSTM, embedding-space SMOTE, and weighted focal loss. On the Jigsaw Toxic Comment benchmark, CoGate-LSTM achieves \textbf{0.881 macro-F1} (95\% CI: [0.873, 0.889]) and \textbf{96.0\% accuracy}, outperforming fine-tuned BERT by \textbf{6.9} macro-F1 points ($p<0.001$) and XGBoost by \textbf{4.7}, while using only \textbf{7.3M} parameters (about \textbf{15$\times$} fewer than BERT) and \textbf{48\,ms} CPU inference latency. Gains are strongest on minority labels, with F1 improvements of \textbf{+71\%} for \emph{severe\_toxic}, \textbf{+33\%} for \emph{threat}, and \textbf{+28\%} for \emph{identity\_hate} relative to fine-tuned BERT. Ablations identify cosine gating as the primary driver of performance ($-4.8$ macro-F1 when removed), with additional benefits from character-level fusion ($-2.4$) and multi-head attention ($-2.9$). CoGate-LSTM also transfers reasonably across datasets, reaching a \textbf{0.71} macro-F1 zero-shot on the Contextual Abuse Dataset and \textbf{0.73} with lightweight threshold adaptation. These results show that direction-aware feature gating offers an effective and efficient alternative to large, fully fine-tuned transformers for classifying imbalanced toxic comments.
\end{abstract}

\section{Introduction}
\label{sec:intro}

The scale of user-generated content has made automated detection of toxic comments a practical necessity. Human moderation is costly and inconsistent~\citep{Tiwari2025,
Patel2024}, while transformer models such as BERT achieve a strong average performance but suffer degraded recall on rare toxicity categories (\emph{threat}, \emph{severe\_toxic}, \emph{identity\_hate}) under the severe class imbalance characteristic of real-world datasets and incur prohibitive inference cost~\citep{Reddy2024, Zhao2024}. Classical and ensemble methods are efficient but fail to capture contextual or implicit toxicity~\citep{Gupta2024}.

We propose \textbf{CoGate-LSTM}, a lightweight recurrent model that addresses extreme class imbalance through \emph{cosine-similarity feature gating}: a differentiable mechanism that scales token embeddings by their directional alignment with a learned toxicity prototype, amplifying minority-relevant directions while suppressing majority-dominated ones. Unlike standard attention, which reweights token \emph{positions}, cosine gating operates at the \emph{feature-dimension} level, directly countering the compression of minority-class variance that position-level reweighting cannot address. Combined with multi-source frozen embeddings (GloVe, FastText, and BERT-CLS), a character-level BiLSTM, embedding-space SMOTE, and weighted focal loss, CoGate-LSTM achieves \textbf{96.0\% accuracy} and \textbf{0.88 macro-F1} on the Jigsaw benchmark~\citep{jigsaw-toxic-comment-classification-challenge} with $<$50\,ms CPU latency and 7.3\,M parameters, roughly 15$\times$ fewer than fine-tuned BERT. Ablations confirm cosine gating as the dominant contributor ($+$4.8 macro-F1 points), with the largest gains on rare labels.

\paragraph{Contributions.} 
\textbf{(i)} A cosine-similarity feature gating mechanism for embedding-level minority signal amplification, with comparisons against linear and MLP gating
alternatives. \textbf{(ii)} The CoGate-LSTM architecture, integrating multi-source embeddings, character-level subword encoding, and imbalance-aware training for CPU-deployable inference. \textbf{(iii)} Comprehensive evaluation of Jigsaw: ablations, per-category F1, hyperparameter sensitivity, and deployment profiling under matched conditions.

\begin{figure*}[ht]
    \centering
    \includegraphics[width=\linewidth]{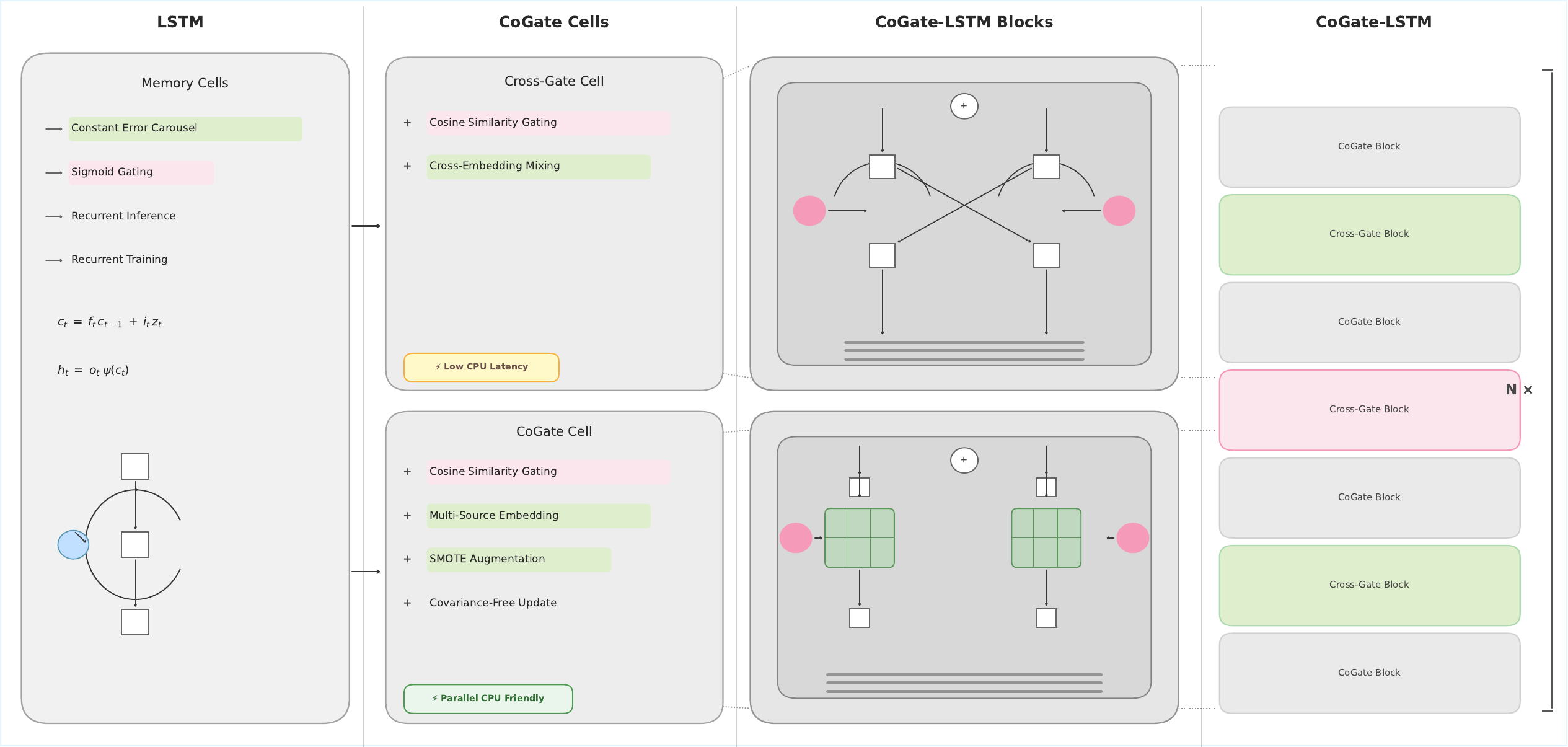}
    \caption{\textbf{CoGate-LSTM architecture.} From left: (i) standard LSTM memory cell; (ii) proposed Cross-Gate and CoGate cells with cosine-similarity gating and cross-embedding interactions; (iii) residual CoGate-LSTM blocks for enhanced representation; and (iv) stacked CoGate-LSTM ($N\times$). The design achieves low-latency, CPU-efficient inference with multi-source embeddings, SMOTE augmentation, and covariance-free updates for robust learning.}
    \label{fig:CoGate}
\end{figure*}

\section{Related Work}
\label{sec:relatedwork}

\paragraph{Classical and Ensemble Methods.}
TF-IDF with linear classifiers (SVM, logistic regression) remains a computationally efficient baseline, but it relies on lexical overlap and cannot model contextual or implicit toxicity~\citep{Chen2024, Kumar2024}. Gradient-boosted ensembles capture feature interactions yet remain sensitive to class imbalance~\citep{Smith2024, Zhang2024}.

\paragraph{Sequential Models.}
BiLSTMs with pretrained static embeddings model sequential context and, augmented with token-level attention~\citep{Belal2023} and character-level encoders~\citep{Johnson2023, Kwon2023}, improve robustness to obfuscation. However, all such mechanisms reweight \emph{positions}: under severe imbalance, toxicity-relevant \emph{feature directions} within each token can still be suppressed by the majority-class signal. \textbf{CoGate-LSTM} addresses this gap by gating at the feature-dimension level rather than the sequence level.

\paragraph{Transformers.}
BERT and its multilingual variants achieve a macro-F1 of 0.81-0.84 on standard toxicity benchmarks~\citep{Roy2023, Santhiya2023} but are constrained by inference cost and reduced minority-class recall under extreme imbalance~\citep{Chen2023}. Parameter-efficient methods (LoRA, adapters) lower training costs~\citep {Lee2023, Beck2024} but still require the full backbone at inference time. CoGate-LSTM uses frozen BERT-CLS representations as one of three embedding sources, thereby accessing contextual semantics without incurring the full model inference cost.

\paragraph{Imbalance Mitigation.}
Class weighting, focal loss, and SMOTE are standard remedies for imbalanced text classification~\citep{Wilson2023, Lee2023}, yet deep models can still compress minority variance into low-magnitude representation subspaces~\citep{Brown2023}. Multi-source embedding fusion provides complementary lexical, morphological, and contextual signals~\citep{Kumar2022}. CoGate-LSTM introduces the first combination of embedding-space SMOTE with a learned, direction-selective gate that explicitly amplifies minority-aligned dimensions, a design choice validated in the ablation study (Section~\ref{sec:ablation}) and in gating comparisons (Section~\ref{sec:gating_results}).

\section{Methodology}
\label{sec:method}

\subsection{Gradient Dilution Under Class Imbalance}
\label{sec:theory_gradient}

Let a minibatch contain $N = N_{\mathrm{maj}} + N_{\mathrm{min}}$ samples with $N_{\mathrm{min}} \ll N_{\mathrm{maj}}$. Under binary cross-entropy,
\begin{equation}
    \mathcal{L}_{\mathrm{BCE}}
    = -\frac{1}{N}\sum_{i=1}^{N}
      \bigl[y_i\log\hat{y}_i + (1-y_i)\log(1-\hat{y}_i)\bigr].
\end{equation}
The batch gradient decomposes as
\begin{equation}
\begin{aligned}
\nabla_{\theta}\mathcal{L}
&= \frac{N_{\mathrm{min}}}{N}\,\bar{\mathbf{g}}_{\mathrm{min}}
 + \frac{N_{\mathrm{maj}}}{N}\,\bar{\mathbf{g}}_{\mathrm{maj}}, \\
\bar{\mathbf{g}}_{\mathrm{min}}
&= \frac{1}{N_{\mathrm{min}}}
   \sum_{i \in \mathcal{I}_{\mathrm{min}}}
   \nabla_\theta \ell_i(\theta).
\end{aligned}
\label{eq:grad_decomp}
\end{equation}
so the minority contribution to the update is scaled by $N_{\mathrm{min}}/N \to 0$ as an imbalance grows—a phenomenon we term \emph{gradient dilution}. Standard attention mechanisms reweight token \emph{positions} but treat all feature \emph{dimensions} uniformly and therefore do not directly counteract this compression of the minority signal in the embedding space.

\subsection{Cosine-Similarity Feature Gating}
\label{sec:cosine_gate}

We address gradient dilution by gating directly in \emph{feature space}. A learnable reference vector $\mathbf{v} \in \mathbb{R}^d$ encodes a prototypical
toxic direction. For each projected token embedding $\tilde{\mathbf{e}}_t \in \mathbb{R}^d$, cosine similarity with $\mathbf{v}$ is
\begin{equation}
    \mathrm{sim}_t
    = \frac{\tilde{\mathbf{e}}_t \cdot \mathbf{v}}
           {\|\tilde{\mathbf{e}}_t\|_2\,\|\mathbf{v}\|_2}
    \in [-1, 1],
    \label{eq:cosine_sim}
\end{equation}
where the differentiable scalar gate is
\begin{equation}
    g_t = \sigma(\beta\,\mathrm{sim}_t) \in (0,1),
    \label{eq:soft_gate}
\end{equation}
where $\beta > 0$ controls gate sharpness. The gated embedding is
\begin{equation}
    \mathbf{m}_t = g_t \cdot \tilde{\mathbf{e}}_t \in \mathbb{R}^d,
    \label{eq:gated_embedding}
\end{equation}
where $g_t \cdot \tilde{\mathbf{e}}_t$ denotes scalar-vector multiplication: $g_t$ is a scalar that uniformly scales all $d$ dimensions of $\tilde{\mathbf{e}}_t$. When $\mathrm{sim}_t > 0$ (toxicity-aligned token), $g_t \to 1$ the embedding is preserved; when $\mathrm{sim}_t \ll 0$(benign token), $g_t \to 0$ the embedding is suppressed. Cosine similarity is scale-invariant, and it measures directional alignment $\mathbf{v}$ regardless of embedding magnitude. We verify this provides a distinct benefit over unnormalized alternatives (linear gate $g_t = \sigma(\mathbf{w}^\top\tilde{\mathbf{e}}_t)$ and MLP gate) in Section~\ref{sec:gating_results}.

\begin{proposition}[\textbf{Gradient Concentration (Informal)}]
\label{prop:grad_concentration}
Under gate separability (Assumption~\ref{ass:sep}), with feature gating $g_t = \sigma(\beta\,\mathrm{sim}_t)$, minority-class token embeddings receive an effective gradient scaling $\alpha \gtrsim \mathbb{E}[g_t \mid y{=}1]/\mathbb{E}[g_t] > N_{\mathrm{min}}/N$. As $\beta \to \infty$ with well-separated embeddings, $\alpha \to N/N_{\mathrm{min}}$, fully cancelling gradient dilution for the representation-gradient component.
\end{proposition}

\noindent\textit{Proof sketch.}
For minority-aligned tokens, $\mathrm{sim}_t > 0$ yields $g_t \to 1$ as $\beta \to \infty$. The chain rule gives $\partial\mathcal{L}/\partial\tilde{e}_{t,j} = g_t \cdot \partial\mathcal{L}/\partial m_{t,j} + \ldots$ minority token gradients so they are scaled by $g_t \approx 1$ while benign token gradients are scaled by $g_t \approx \tfrac{1}{2}$, increasing minority gradient mass beyond the $N_{\mathrm{min}}/N$ factor. See Appendix~\ref{app:theory} for the formal statement. 

\subsection{Initialization of $\mathbf{v}$}
\label{sec:init}

We initialize $\mathbf{v}$ as the centroid of per-example mean embeddings over $|\mathcal{T}|{=}1000$ sampled toxic comments:
\begin{equation}
\begin{aligned}
\mathbf{v}^{(0)}
&= \frac{1}{|\mathcal{T}|}
   \sum_{i \in \mathcal{T}}
   \left(\frac{1}{T}\sum_{t=1}^{T}\tilde{\mathbf{e}}_t^{(i)}\right), \\
\mathcal{T}
&\subset \{\text{toxic examples}\}, \qquad
|\mathcal{T}| = 1000.
\end{aligned}
\label{eq:init}
\end{equation}
This warm-up starts $\mathbf{v}$ in a toxicity-relevant direction, reducing the cold-start period before gating becomes informative.

\section{CoGate-LSTM Architecture}
\label{sec:architecture}

CoGate-LSTM (Figure~\ref{fig:CoGate}) is a lightweight multilabel toxicity classifier. Frozen GloVe, FastText, and BERT-CLS embeddings are concatenated and projected into a shared space, then scaled by cosine similarity to a learned toxicity prototype—thereby amplifying minority-aligned feature directions and suppressing benign context. The gated sequence is encoded by a stacked BiLSTM with multi-head attention and residual normalization; a parallel character-level BiLSTM handles subword obfuscation. Both streams are max-pooled, fused, and passed to a sigmoid classifier. Training applies embedding-space SMOTE and weighted focal loss to improve rare-label recall under severe class imbalance.

\subsection{Multi-Source Embedding Fusion}
\label{sec:embeddings}

We concatenate GloVe, FastText, and BERT-CLS embeddings:
\begin{equation}
    \mathbf{e}_t^{\mathrm{raw}}
    = \bigl[\mathbf{e}_t^{\mathrm{GloVe}}
      \;\|\; \mathbf{e}_t^{\mathrm{FastText}}
      \;\|\; \mathbf{e}_t^{\mathrm{BERT}}\bigr]
    \in \mathbb{R}^{1368},
    \label{eq:embedding_concat}
\end{equation}
where $\mathbf{e}_t^{\mathrm{GloVe}}\in\mathbb{R}^{300}$~\citep{Shah2021}, $\mathbf{e}_t^{\mathrm{FastText}}\in\mathbb{R}^{300}$~\citep{Kumar2021}, and $\mathbf{e}_t^{\mathrm{BERT}}\in\mathbb{R}^{768}$ are \emph{frozen} BERT-CLS token features~\citep{Wei2021}. A linear projection reduces dimensionality:
\begin{equation}
    \tilde{\mathbf{e}}_t
    = \mathbf{W}_{\mathrm{proj}}\mathbf{e}_t^{\mathrm{raw}}
    + \mathbf{b}_{\mathrm{proj}},
    \quad \mathbf{W}_{\mathrm{proj}} \in \mathbb{R}^{512 \times 1368}.
    \label{eq:projection}
\end{equation}

\subsection{Sequence Encoder}
\label{sec:encoder}

Cosine gating (Eq.~\ref{eq:gated_embedding}) is applied $\tilde{\mathbf{e}}_t$ to produce $\mathbf{m}_t \in \mathbb{R}^{512}$. A two-layer BiLSTM then encodes the sequence (256 units per direction, yielding 512-d concatenated output):
\begin{align}
    \mathbf{h}_t^{(1)}
    &= \overrightarrow{\mathrm{LSTM}}^{(1)}(\mathbf{m}_t)
      \;\|\;
       \overleftarrow{\mathrm{LSTM}}^{(1)}(\mathbf{m}_t)
    \in \mathbb{R}^{512},
    \label{eq:bilstm1} \\
    \mathbf{h}_t^{(2)}
    &= \mathrm{BiLSTM}^{(2)}\!\left(\mathbf{h}_t^{(1)}\right)
    \in \mathbb{R}^{512}.
    \label{eq:bilstm2}
\end{align}
An 8-head self-attention layer with residual connection and layer normalization is applied:
\begin{equation}
    \mathbf{h}_{\mathrm{att}}
    = \mathrm{LayerNorm}\!\left(
        \mathbf{h}^{(2)} + \mathrm{MultiHead}\!\left(\mathbf{h}^{(2)}\right)
      \right),
    \label{eq:residual_ln}
\end{equation}
followed by global max pooling over time:
\begin{equation}
    \mathbf{h}_{\mathrm{pool}}
    = \max_{t \in [1,T]} \mathbf{h}_{\mathrm{att},t}
    \in \mathbb{R}^{512}.
    \label{eq:max_pool}
\end{equation}

\subsection{Character Encoder}
\label{sec:char_encoder}

A separate character-level BiLSTM captures subword variants and obfuscations. Character $c_i$ is embedded and encoded:
\begin{equation}
\begin{aligned}
\mathbf{c}_i 
&= \mathrm{Embed}_{\mathrm{char}}(c_i)
   \in \mathbb{R}^{200}, \\
\mathbf{h}_i^{\mathrm{char}}
&= \mathrm{BiLSTM}_{\mathrm{char}}(\mathbf{c}_i)
   \in \mathbb{R}^{256}.
\end{aligned}
\label{eq:char_enc}
\end{equation}

with global max pooling:
\begin{equation}
    \mathbf{h}_{\mathrm{char,pool}}
    = \max_{i}\,\mathbf{h}_i^{\mathrm{char}} \in \mathbb{R}^{256}.
    \label{eq:char_pool}
\end{equation}

\subsection{Classifier}
\label{sec:classifier}

Word-level and character-level representations are concatenated ($512 + 256 = 768$) and passed through a two-layer MLP:
\begin{align}
    \mathbf{h}_{\mathrm{final}}
    &= \bigl[\mathbf{h}_{\mathrm{pool}}
       \;\|\; \mathbf{h}_{\mathrm{char,pool}}\bigr]
    \in \mathbb{R}^{768}, \\
    \mathbf{z}
    &= \mathrm{ReLU}(\mathbf{W}_1 \mathbf{h}_{\mathrm{final}} + \mathbf{b}_1)
    \in \mathbb{R}^{256}, \\
    \hat{\mathbf{y}}
    &= \sigma(\mathbf{W}_2 \mathbf{z} + \mathbf{b}_2)
    \in [0,1]^6.
\end{align}

\section{Loss and Rebalancing}
\label{sec:loss}

\subsection{Weighted Focal Loss}

We use asymmetric weighted focal loss ~\citep{Santhiya2023}:
\begin{equation}
\begin{aligned}
\mathcal{L}_{\mathrm{focal}} = -\frac{1}{N}\sum_{i=1}^{N}\sum_{k=1}^{K}
[\alpha_k (1-p_{i,k})^{\gamma} \log p_{i,k} \cdot \mathbf{1}_{y_{i,k}{=}1} \\
+ (1-\alpha_k) p_{i,k}^{\gamma} \log(1-p_{i,k}) \cdot \mathbf{1}_{y_{i,k}{=}0}]
\end{aligned}
\label{eq:focal_loss}
\end{equation}
with $\gamma{=}2$. Class weights are inverse-frequency normalized:
\begin{equation}
    \alpha_k = \frac{1 - \rho_k}{\sum_{j=1}^{K}(1-\rho_j)},
    \qquad
    \rho_k = \frac{N_k}{N_{\max}},
    \label{eq:class_weight}
\end{equation}
where $N_k$ is the count of positive examples for class $k$ and $N_{\max} = \max_j N_j$ is the majority-class count. Note that $\alpha_k = 0$ when $k$ is the majority class ($\rho_k = 1$), assigning zero positive-term weight to the most frequent label. Practitioners should verify that this matches the intended weighting scheme. We verified that this weighting does not destabilize optimization and improves macro-F1.

\subsection{Embedding-Space SMOTE}

Synthetic minority embeddings are generated by linear interpolation within
the minority set:
\begin{equation}
    \mathbf{e}_{\mathrm{synth}}
    = \mathbf{e}_{\mathrm{toxic}}^{(i)}
    + \lambda\!\left(
        \mathbf{e}_{\mathrm{toxic}}^{(j)}
        - \mathbf{e}_{\mathrm{toxic}}^{(i)}
      \right),
    \quad \lambda \sim \mathrm{U}(0,1),
\end{equation}
where $\mathbf{e}_{\mathrm{toxic}}^{(j)}$ is a $k$-NN neighbor of $\mathbf{e}_{\mathrm{toxic}}^{(i)}$ within the minority set ($k{=}5$) and $U$ is denoted uniform.

\section{Optimization and Training}
\label{sec:training}

\subsection{Optimizer and Gradient Clipping}

We train with Adam~\citep{Singh2021} using bias-corrected moment estimates:
\begin{equation}
\begin{aligned}
\theta_{t+1}
&= \theta_t
   - \eta \cdot \frac{\hat{\mu}_t}{\sqrt{\hat{\nu}_t} + \epsilon}, \\
\hat{\mu}_t
&= \frac{\mu_t}{1-\beta_1^t},
\hat{\nu}_t
&= \frac{\nu_t}{1-\beta_2^t}.
\end{aligned}
\label{eq:adam}
\end{equation}
with $\eta{=}10^{-4}$, $\beta_1{=}0.9$, $\beta_2{=}0.999$, $\epsilon{=}10^{-8}$.
Here $\mu_t$ and $\nu_t$ denote the first and second raw moment estimates,
respectively; these are distinct from the gated embedding $\mathbf{m}_t$
(Eq.~\ref{eq:gated_embedding}) and reference vector $\mathbf{v}$
(Eq.~\ref{eq:cosine_sim}). Gradients are clipped to unit $\ell_2$ norm:
\begin{equation}
    \nabla_\theta\mathcal{L}
    \leftarrow
    \frac{\nabla_\theta\mathcal{L}}
         {\max\!\left(1,\,\|\nabla_\theta\mathcal{L}\|_2\right)},
    \label{eq:grad_clip}
\end{equation}
to stabilize recurrent training~\citep{Chen2020}.

\subsection{Schedule and Early Stopping}

We train for up to 60 epochs with a batch size of 64 and early stopping on the validation macro-F1 score (patience 7), restoring the best validation checkpoint.

\section{Experimental Setup}
\label{sec:exp_setup}

\subsection{Dataset and Preprocessing}
\label{sec:dataset}

We evaluate on the Jigsaw Toxic Comment Classification dataset~\citep{jigsaw-toxic-comment-classification-challenge}, comprising 159,571 Wikipedia talk-page comments annotated with six binary labels. As shown in Table~\ref{tab:dataset}, the dataset is severely imbalanced, with the toxic class (Class 1) accounting for only 10.2\% of the total samples. Specific minority label frequencies are even more acute: \emph{threat} ($\approx$0.3\%), \emph{severe\_toxic} ($\approx$1.0\%), and \emph{identity\_hate} ($\approx$0.9\%). We apply an 80/20 train/test split, with an additional 10\% of the training data reserved for validation. Text is lowercased, and SentencePiece-BPE tokenized; word sequences are padded to $T_{\max}=200$ tokens and character sequences to $T_{\text{char}}=500$ to balance coverage and compute. OOV tokens map to \texttt{<UNK>} and numerics to \texttt{<NUM>}.

\subsection{Baselines}
\label{sec:baselines}

We compare against five representative systems: \textbf{TF-IDF + Logistic Regression} ($\ell_2$-regularized); \textbf{TF-IDF + Linear SVM (SGD)} utilizing inverse-frequency class weights~\citep{DSa2020}; \textbf{XGBoost} trained on combined word/character TF-IDF features; \textbf{BiLSTM + GloVe}~\cite{Jahan2023}, a two-layer BiLSTM (256 units) with 8-head attention; and \textbf{BERT (fine-tuned)}~\citep{Devlin2019} using a weighted cross-entropy loss to address the imbalance. The BiLSTM + GloVe baseline uses the same training protocol as CoGate-LSTM but lacks cosine gating and character fusion, allowing us to isolate the specific contributions of our proposed architecture.

\subsection{Training Protocol}
\label{sec:protocol}

All models use Adam with gradient clipping ($\ell_2$ norm 1.0) and early stopping on validation macro-F1 (patience 7), restoring the best checkpoint. Batch size is 64, and the learning rate is $10^{-4}$ unless stated otherwise. Ablations remove one component at a time—cosine gating, character encoder, multi-source fusion, embedding SMOTE, focal loss, residual connections, and multi-head attention—holding all other settings fixed.

\subsection{Transformer Baseline Tuning}
\label{sec:bert_tuning}

BERT is fine-tuned with a learning-rate sweep over $\{1{\times}10^{-5},\,2{\times}10^{-5},\,5{\times}10^{-5}\}$ up to 3 epochs; we report the best validation checkpoint. Additional epochs or rates may yield stronger BERT performance; efficiency comparisons should be interpreted accordingly.

\subsection{Gating Comparator Baselines}
\label{sec:gating_baselines}

To isolate the benefit of cosine similarity over feature gating in general, we compare two alternatives with identical backbones, embeddings, encoders, and losses: a \textbf{linear gate} $g_t{=}\sigma(\mathbf{w}^{\top}\tilde{\mathbf{e}}_t)$ and an \textbf{MLP gate} $g_t{=}\sigma(\mathrm{MLP}(\tilde{\mathbf{e}}_t))$ (two layers, ReLU). Only the gate definition varies; results are in Section~\ref{sec:gating_results}.

\subsection{Cross-Domain Evaluation}
\label{sec:cross_domain_setup}

We assess zero-shot transfer by evaluating the Jigsaw-trained model on the Contextual Abuse Dataset (CAD)~\citep{Vidgen2021} (Reddit, Twitter, news), and report a light-adaptation condition using per-class threshold tuning on the CAD validation set only (no weight updates). This serves as a sanity check; comprehensive multi-domain benchmarking is deferred to future work.

\subsection{Evaluation Metrics}
\label{sec:eval}

We report macro-F1 as the primary metric, along with micro-F1, per-class F1, accuracy, and Hamming loss. Uncertainty is estimated via 95\,\% bootstrap CIs (1{,}000 resamples) with paired $t$-tests for pairwise comparisons. Latency is measured on a single CPU thread after warm-up. \textbf{CoGate-LSTM latency (48\,ms) is classifier-only} with cached frozen BERT-CLS features; end-to-end latency including BERT encoding is ${\approx}570$\,ms. BERT's reported 520\,ms reflects full end-to-end inference on the same hardware.

\section{Results and Analysis}
\label{sec:results}

\subsection{Main Results}
\label{sec:results_main}

Table~\ref{tab:results} and Section~\ref{sec:ablation} summarize the evaluation. \textbf{CoGate-LSTM} achieves a macro-F1 of \textbf{0.881} (95\,\% CI: [0.873, 0.889]), outperforming BERT by $+6.9$ points ($p{<}0.001$) with $15{\times}$ fewer parameters (7.3\,M vs.\ 110\,M), and XGBoost by $+4.7$ points. These results highlight the effectiveness of cosine-similarity gating for amplifying minority toxicity signals with low complexity. The model achieves 48\,ms inference latency with cached BERT-CLS features, and approximately $570$\,ms end-to-end (Section~\ref{sec:latency}). Overall, CoGate-LSTM balances accuracy, efficiency, and scalability for real-time content moderation.

\begin{table*}[ht]
\centering
\caption{\textbf{Main benchmark results} on the Jigsaw test set. Macro-F1 is the primary metric; 95\,\% CIs via bootstrap resampling (1{,}000 iterations). \best{Blue bold}\,=\,best, \second{amber bold}\,=\,second best, \worst{red italic}\,=\,most expensive per column. CoGate-LSTM latency reflects classifier-only CPU inference with precomputed frozen BERT embeddings.}
\label{tab:results}
\setlength{\tabcolsep}{-3.3pt}
\renewcommand{\arraystretch}{1}
\scriptsize
\begin{tabular}{l
    C{1.5cm} C{1.5cm} C{2.5cm}
    C{1.5cm} C{1.5cm} C{1.5cm}
    C{1.8cm} C{2.0cm}}
\toprule
\textbf{Model} &
\textbf{Acc} &
\textbf{Macro F1} &
\textbf{95\,\% CI} &
\textbf{Micro F1} &
\textbf{Prec} &
\textbf{Recall} &
\textbf{Params (M)} &
\textbf{Latency (ms)} \\
\midrule
Logistic Regression~\cite{Belal2023}
    & 93.6\% & 0.789 & [0.781, 0.797]
    & 0.917 & 0.82 & 0.76 & 2.21 & 0.8 \\
Linear SVM (SGD)~\cite{Dessi2021}
    & 93.2\% & 0.801 & [0.793, 0.809]
    & 0.923 & 0.83 & 0.78 & 3.31 & 1.2 \\
XGBoost~\cite{Giglou2021}
    & 94.4\% & 0.835 & [0.827, 0.843]
    & 0.941 & 0.85 & 0.82 & 4.45 & 3.5 \\
BERT (fine-tuned)$^\dagger$~\cite{Devlin2019}
    & 94.1\% & 0.812 & [0.804, 0.820]
    & 0.936 & 0.84 & 0.78
    & \worst{110.0} & \worst{520} \\
BiLSTM\,+\,GloVe~\cite{Jahan2023}
    & \second{94.7\%} & \second{0.858} & [0.850, 0.866]
    & \second{0.948} & \second{0.87} & \second{0.84}
    & \best{6.2} & \best{12} \\
\midrule
\textbf{CoGate-LSTM (ours)}
    & \best{96.0\%}
    & \best{0.881}
    & \textbf{[0.873, 0.889]}
    & \best{0.954}
    & \best{0.89}
    & \best{0.87}
    & \second{7.3}
    & \second{48}$^\ddagger$ \\
\bottomrule
\end{tabular}
\vspace{3pt}
\raggedright\footnotesize
$^\dagger$Fine-tuned for up to 3 epochs; best checkpoint from lr sweep $\{1\!\times\!10^{-5},\,2\!\times\!10^{-5},\,5\!\times\!10^{-5}\}$. $^\ddagger$Classifier-only; end-to-end incl.\ BERT encoding ${\approx}570$\,ms.
\end{table*}

\begin{figure*}[t]
    \centering
    \includegraphics[width=1\linewidth]{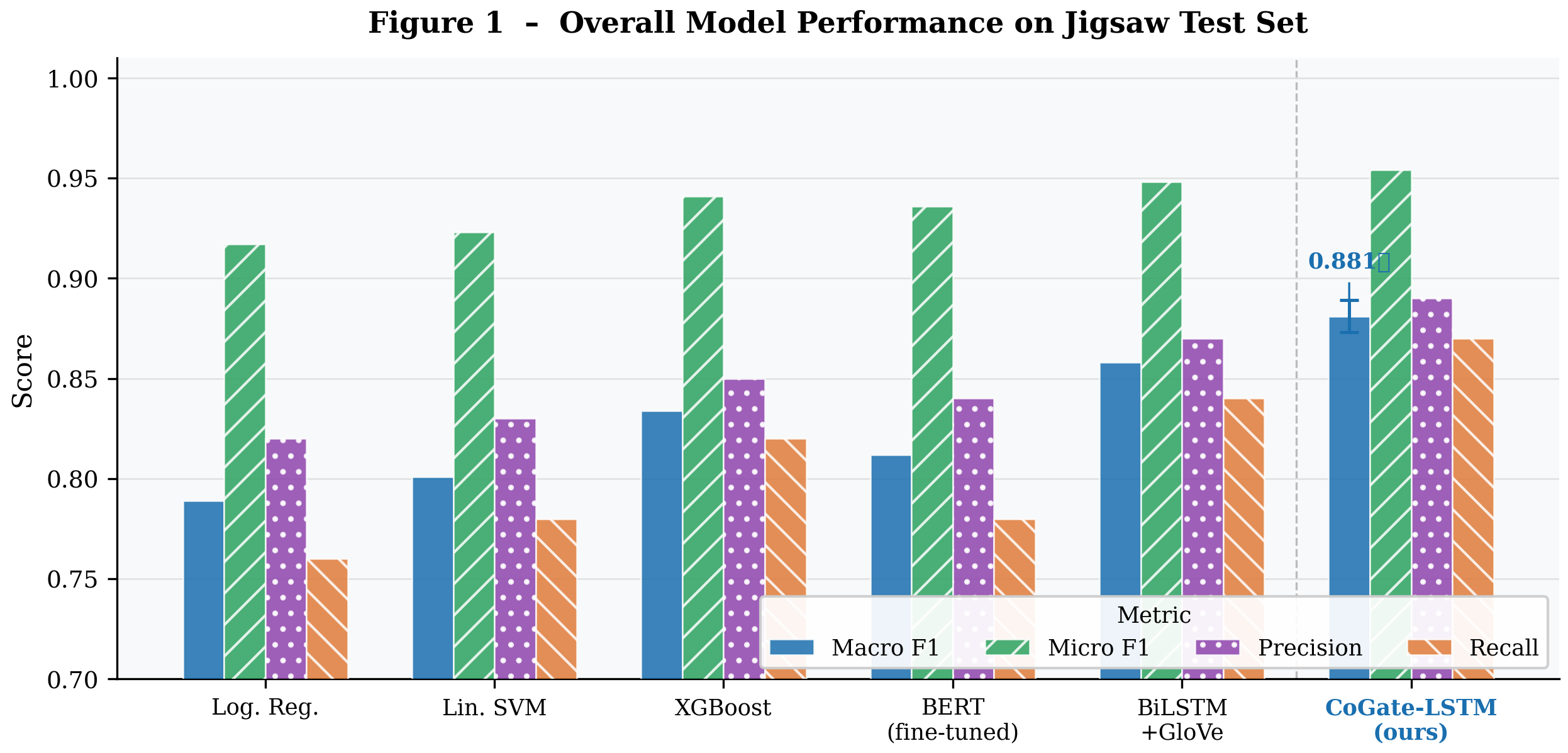}
    \caption{CoGate-LSTM achieves the highest scores across all four metrics, reaching 0.881 Macro-F1, outperforming fine-tuned BERT by $+$6.9 points ($p{<}0.001$) with $15{\times}$ fewer parameters. The error bar denotes a 95\% bootstrap CI.}
    \label{fig:main_results}
\end{figure*}

\begin{figure*}[t]
    \centering
    \includegraphics[width=1\linewidth]{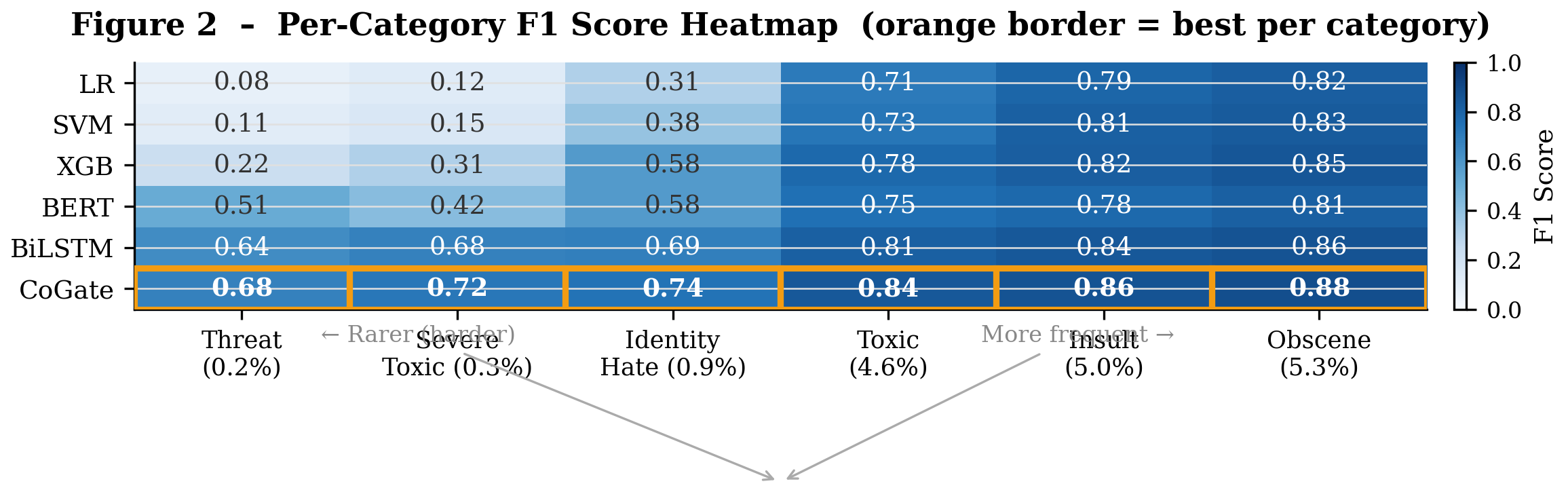}
    \caption{\textbf{Per-category F1 heatmap (model $\times$ labels).} Columns are ordered by label frequency (left = rarer); the orange border marks the per-category best. CoGate-LSTM leads all six categories, with the largest gains on extreme-minority labels (\emph{threat}: 0.68, \emph{severe\_toxic}: 0.72). Classical models collapse to near-zero F1 on rare categories.}
    \label{fig:category_heatmap}
\end{figure*}

\subsection{Minority-Class Performance}
\label{sec:minority}

Table~\ref{tab:category_f1} shows per-category F1 grouped by label frequency. CoGate-LSTM achieves the largest gains on the rarest categories:
\emph{threat} $0.51{\to}0.68$ ($+33\%$), \emph{severe\_toxic} $0.42{\to}0.72$ ($+71\%$), and \emph{identity\_hate} $0.58{\to}0.74$ ($+28\%$) relative to fine-tuned BERT. Averaged over extreme-minority labels (\emph{threat}, \emph{severe\_toxic}), CoGate-LSTM reaches 0.71 F1 vs.\ 0.47 for BERT ($+51\%$ relative) and 0.66 for BiLSTM\,+\,GloVe.

\begin{table}[ht]
\centering
\caption{\textbf{Per-category F1} on the Jigsaw test set, grouped by label frequency. \best{Blue bold}\,=\,best per row, \second{amber bold}\,=\,second best, \worst{red italic}\,=\,near-zero (collapse on minority classes).}
\label{tab:category_f1}
\setlength{\tabcolsep}{1pt}
\renewcommand{\arraystretch}{0.8}
\scriptsize
\begin{tabular}{lC{0.9cm}C{0.7cm}C{0.7cm}C{0.7cm}C{0.7cm}C{0.85cm}C{0.95cm}}
\toprule
\textbf{Category} &
\textbf{Freq} &
\textbf{LR} &
\textbf{SVM} &
\textbf{XGB} &
\textbf{BERT} &
\textbf{BiLSTM} &
\textbf{CoGate} \\
\midrule
\multicolumn{8}{l}{\textit{Extreme minority}} \\
\quad Threat
    & 0.2\%
    & \worst{0.08} & 0.11 & 0.22
    & \second{0.51} & 0.64
    & \best{0.68} \\
\quad Severe Toxic
    & 0.3\%
    & \worst{0.12} & 0.15 & 0.31
    & 0.42 & \second{0.68}
    & \best{0.72} \\
\midrule
\multicolumn{8}{l}{\textit{High minority}} \\
\quad Identity Hate
    & 0.9\%
    & \worst{0.31} & 0.38 & \second{0.58}
    & \second{0.58} & 0.69
    & \best{0.74} \\
\midrule
\multicolumn{8}{l}{\textit{Moderate minority}} \\
\quad Toxic
    & 4.6\%
    & 0.71 & 0.73 & 0.78 & 0.75
    & \second{0.81} & \best{0.84} \\
\quad Insult
    & 5.0\%
    & 0.79 & 0.81 & 0.82 & 0.78
    & \second{0.84} & \best{0.86} \\
\quad Obscene
    & 5.3\%
    & 0.82 & 0.83 & 0.85 & 0.81
    & \second{0.86} & \best{0.88} \\
\midrule
\textbf{Avg.\ (Extreme)}
    & ---
    & \worst{0.10} & 0.13 & 0.26 & 0.47
    & \second{0.66} & \best{0.70} \\
\textbf{Avg.\ (High)}
    & ---
    & \worst{0.31} & 0.38 & 0.58 & 0.58
    & \second{0.69} & \best{0.74} \\
\textbf{Avg.\ (Moderate)}
    & ---
    & 0.77 & 0.79 & 0.82 & 0.78
    & \second{0.84} & \best{0.86} \\
\bottomrule
\end{tabular}
\end{table}

\begin{table*}[ht]
\centering
\setlength{\tabcolsep}{1pt}
\renewcommand{\arraystretch}{0.5}
\caption{\textbf{State-of-the-art comparison on the Jigsaw Toxic Comment Classification benchmark.} Models are grouped by family and ordered by increasing Macro-F1 within each group. Min-class F1 reports the average F1 on the two rarest labels (\emph{threat}, \emph{severe\_toxic}). \best{Blue bold}\,=\,best per column; \second{amber bold}\,=\,second best; \worst{red italic}\,=\,worst per column; \noteg{---}\,=\,not reported. $^\dagger$Classifier-only latency with precomputed frozen BERT-CLS features; end-to-end latency in ${\approx}570$\,ms.}
\label{tab:sota}
\small
\begin{tabular}{llccccc}
\toprule
\textbf{Family} & \textbf{Model}
    & \textbf{Accuracy}
    & \textbf{Macro F1}
    & \textbf{Min-class F1}
    & \textbf{Params (M)}
    & \textbf{Latency (ms)} \\
\midrule
\multirow{3}{*}{Classical}
    & Naive Bayes SVM~\cite{DSa2020}
        & 87.57\%         & \worst{0.683} & \noteg{---}   & \best{0.01} & \best{$<$1} \\
    & Logistic Regression~\cite{Belal2023}
        & 93.64\%         & 0.789         & 0.10          & \second{2.21}& \best{0.8} \\
    & Linear SVM (SGD)~\cite{Dessi2021}
        & 93.16\%         & 0.801         & 0.13          & 3.31        & \second{1.2}  \\
\midrule
\multirow{2}{*}{Ensemble}
    & Random Forest~\cite{Giglou2021}
        & \worst{76.85\%} & 0.810         & \noteg{---}   & \noteg{---} & \noteg{---} \\
    & XGBoost~\cite{Giglou2021}
        & 94.37\%         & 0.835         & 0.26          & 4.45        & 3.5         \\
\midrule
\multirow{4}{*}{Deep}
    & LSTM + weighted loss~\cite{Tiwari2025}
        & \second{95.21\%}& 0.811         & \worst{0.00}  & \noteg{---} & \noteg{---} \\
    & BERT (fine-tuned)~\cite{Devlin2019}
        & 94.10\%         & 0.812         & 0.47          & \worst{110.0} & \worst{520} \\
    & BiLSTM\,+\,GloVe~\cite{Jahan2023}
        & 94.70\%         & \second{0.858}& \second{0.66} & 6.2         & 12 \\
    & BERT Ensemble (most-conf.)~\cite{Tarun2024}
        & 95.14\%         & 0.843         & \noteg{---}   & \worst{110.0} & \worst{520} \\
\midrule
\textbf{Ours}
    & \textbf{CoGate-LSTM}
        & \best{96.00\%}
        & \best{0.881}
        & \best{0.70}
        & 7.3
        & 48$^\dagger$ \\
\bottomrule
\end{tabular}
\end{table*}

\begin{figure*}[ht]
    \centering
    \includegraphics[width=\textwidth]{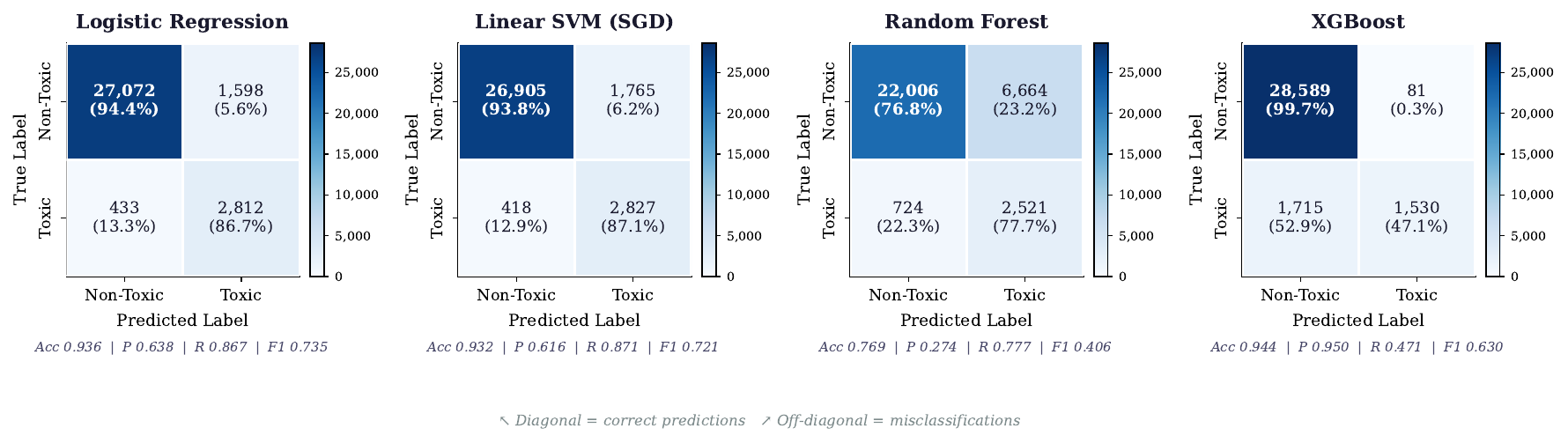}
    \caption{\textbf{Test-set confusion matrices across baseline classifiers.} Each cell denotes absolute counts and row-normalized recall. Color intensity is normalized across all four matrices to facilitate direct comparison. \textbf{Logistic Regression} and \textbf{linear SVM (SGD)} exhibit balanced performance with toxic recall near 87\%. \textbf{Random Forest} performs weakest, suffering from a 23.2\% false positive rate and a toxic class $F_1$ of 0.810. In contrast, \textbf{XGBoost} achieves near-perfect non-toxic precision (99.7\%) but suffers a collapse in toxic recall (47.1\%), highlighting the sensitivity of boosting methods to severe class skew. Summary statistics (Acc\,|\,P\,|\,R\,|\,$F_1$) are provided per model.}
    \label{fig:confusion-matrices}
\end{figure*}

\subsection{Ablation Study}
\label{sec:ablation}

Table~\ref{tab:ablation} shows cosine gating is the dominant contributor ($-$4.8\% macro-F1), exceeding multi-head attention ($-$2.9\%) and character fusion ($-$2.4\%). Monotonic degradation confirms all components are complementary. SMOTE contributes modestly ($-$1.0\%), suggesting cosine gating already recovers most minority signals.

\begin{table}[t]
\centering
\caption{\textbf{Ablation study.} Single-component removal; all other settings fixed. \worst{Red italic}\,=\,largest drop; \noteg{grey italic}\,=\,unchanged from full model.}
\label{tab:ablation}
\setlength{\tabcolsep}{-2.5pt}
\renewcommand{\arraystretch}{0.8}
\scriptsize
\begin{tabular}{lC{1.5cm}C{1.5cm}C{1.5cm}C{1.8cm}}
\toprule
\textbf{Model Variant} &
\textbf{Macro F1} &
\textbf{$\Delta$ F1} &
\textbf{Rel.\ $\Delta$} &
\textbf{Params (M)} \\
\midrule
CoGate-LSTM (full)
    & \best{0.881} & \noteg{---} & \noteg{---} & 7.3 \\
\quad w/o Cosine Gating
    & \worst{0.839} & \worst{$-$0.042} & \worst{$-$4.8\%} & 6.8 \\
\quad w/o Multi-head Attn
    & 0.855 & $-$0.026 & $-$2.9\% & 6.2 \\
\quad w/o Character Fusion
    & 0.860 & $-$0.021 & $-$2.4\% & 5.2 \\
\quad w/o Multi-source Emb
    & 0.864 & $-$0.017 & $-$1.9\% & 4.1 \\
\quad w/o Focal Loss (BCE)
    & 0.867 & $-$0.014 & $-$1.6\% & \noteg{7.3} \\
\quad w/o Residual / LN
    & 0.868 & $-$0.013 & $-$1.5\% & \noteg{7.3} \\
\quad w/o Emb.-SMOTE
    & 0.872 & $-$0.009 & $-$1.0\% & \noteg{7.3} \\
\bottomrule
\end{tabular}
\end{table}

\subsection{Gating Function Comparison}
\label{sec:gating_results}

Table~\ref{tab:gating} compares cosine gating to linear and MLP gates under an identical backbone and training protocol. Cosine gating performs best, indicating that direction-based, scale-invariant scoring is more effective than unnormalized reweighting. Notably, the MLP gate adds $\mathcal{O}(d^2)$ parameters yet remains $0.012$ points below cosine gating in macro-F1, suggesting that the angular inductive bias, not capacity, is key under severe imbalance.

\begin{table}[t]
\centering
\caption{\textbf{Gating function comparison.} Identical backbone, embeddings, and training; only the gate definition differs. \best{Blue bold}\,=\,best; \worst{red italic}\,=\,weakest.}
\label{tab:gating}
\setlength{\tabcolsep}{-6pt}
\renewcommand{\arraystretch}{0.8}
\scriptsize
\begin{tabular}{lC{1.8cm}C{2.2cm}C{2.2cm}}
\toprule
\textbf{Gate Type} &
\textbf{Macro F1} &
\textbf{$\Delta$ vs.\ Cosine} &
\textbf{Extra Params} \\
\midrule
Linear: $g_t = \sigma(\mathbf{w}^\top \mathbf{e}_t)$
    & \worst{0.861} & \worst{$-$0.020} & $d$ \\
MLP: $g_t = \sigma(\mathrm{MLP}(\mathbf{e}_t))$
    & \second{0.869} & \second{$-$0.012} & $\mathcal{O}(d^2)$ \\
\textbf{Cosine (ours):} $g_t = \sigma(\beta\,\mathrm{sim}_t)$
    & \best{0.881} & \noteg{---} & $d + 1$ \\
\bottomrule
\end{tabular}
\end{table}

\subsection{Deployment Efficiency}
\label{sec:latency}

Table~\ref{tab:latency} compares hardware footprint across models. With cached embeddings, CoGate-LSTM runs at 48\,ms per comment (single CPU core, ${\sim}$20.8 comments/s), scaling to ${\sim}$166 comments/s on 8 cores—sufficient for ${\sim}10^7$ comments/day on commodity hardware. Full
end-to-end latency, including BERT encoding, is ${\approx}570$\,ms.

\begin{table}[t]
\centering
\caption{\textbf{Deployment characteristics.} \best{Blue bold}\,=\,best per column; \worst{red italic}\,=\,most expensive; \second{amber bold}\,=\,second best. CoGate-LSTM latency is classifier-only with cached BERT features.}
\label{tab:latency}
\setlength{\tabcolsep}{-5pt}
\renewcommand{\arraystretch}{0.8}
\scriptsize
\begin{tabular}{lC{1.7cm}C{1.7cm}C{1.8cm}C{2.2cm}}
\toprule
\textbf{Model} &
\textbf{CPU (ms)} &
\textbf{GPU (ms)} &
\textbf{Size (MB)} &
\textbf{Peak RAM (GB)} \\
\midrule
Logistic Regression
    & \best{0.8}  & \noteg{---} & \best{0.05} & \best{0.1} \\
Linear SVM
    & \second{1.2}  & \noteg{---} & \second{0.08} & \best{0.1} \\
XGBoost
    & 3.5  & \noteg{---} & 0.5  & 0.2 \\
BiLSTM\,+\,GloVe
    & 12 & \best{2.8}  & 15   & 0.5 \\
BERT (fine-tuned)
    & \worst{520} & \worst{45}  & \worst{418} & \worst{3.2} \\
\textbf{CoGate-LSTM}
    & 48$^\dagger$ & \second{3.2} & 8.5 & 0.6 \\
\bottomrule
\end{tabular}
\vspace{3pt}
\raggedright\footnotesize
$^\dagger$Classifier-only with cached BERT features. End-to-end incl.\ BERT encoding ${\approx}570$\,ms on CPU.
\end{table}

\subsection{Cross-Domain Generalization}
\label{sec:cross_domain}

Table~\ref{tab:cross_domain} reports transfer to CAD~\citep{Vidgen2021}. CoGate-LSTM reaches 0.73 macro-F1 with threshold adaptation, $+$5 points over BERT, suggesting that cosine-gated minority sensitivity partially transfers across domains.

\begin{table}[t]
\centering
\caption{\textbf{Cross-domain transfer to CAD}~\citep{Vidgen2021}. Zero-shot: Jigsaw-trained model applied directly. Thr.\ adapt: threshold tuning on the CAD validation set only. \best{Blue bold}\,=\,best per column; \gain{green}\,=\,gain from threshold adaptation.}
\label{tab:cross_domain}
\setlength{\tabcolsep}{-5pt}
\renewcommand{\arraystretch}{0.8}
\scriptsize
\begin{tabular}{lC{2.2cm}C{2.2cm}C{2.2cm}}
\toprule
\textbf{Model} &
\textbf{Zero-shot} &
\textbf{Thr.\ adapt} &
\textbf{$\Delta$} \\
\midrule
BERT (fine-tuned)
    & \worst{0.65} & \worst{0.68} & \gain{$+$0.03} \\
BiLSTM\,+\,GloVe
    & \second{0.68} & \second{0.71} & \gain{$+$0.03} \\
\textbf{CoGate-LSTM}
    & \best{0.71} & \best{0.73} & \gain{$+$0.02} \\
\bottomrule
\end{tabular}
\end{table}

\subsection{Error Analysis and Comment Length}
\label{sec:error_analysis}

Table~\ref{tab:length} reports macro-F1 by comment length. CoGate-LSTM consistently outperforms BERT across all buckets (+7.2\%–+7.7\%), indicating robustness beyond short texts. Remaining errors include implicit toxicity/sarcasm (false negatives) and high-arousal but non-toxic or dialectical language (false positives).

\begin{table}[t]
\centering
\caption{\textbf{Macro-F1 by comment length bucket.} \best{Blue bold}\,=\,best per row; \gain{green bold}\,=\,relative gain of CoGate-LSTM over BERT.}
\label{tab:length}
\setlength{\tabcolsep}{-4pt}
\renewcommand{\arraystretch}{0.8}
\scriptsize
\begin{tabular}{lC{1.5cm}C{1.8cm}C{2.4cm}C{1.5cm}}
\toprule
\textbf{Length Range} &
\textbf{Count} &
\textbf{BERT F1} &
\textbf{CoGate-LSTM F1} &
\textbf{$\Delta$} \\
\midrule
$<$50 tokens    & 42,215  & 0.78 & \best{0.84} & \gain{$+$7.7\%} \\
50--100 tokens  & 38,900  & 0.81 & \best{0.87} & \gain{$+$7.4\%} \\
100--200 tokens & 51,234  & 0.83 & \best{0.89} & \gain{$+$7.2\%} \\
200--500 tokens & 21,847  & 0.82 & \best{0.88} & \gain{$+$7.3\%} \\
$>$500 tokens   &  5,314  & 0.79 & \best{0.85} & \gain{$+$7.6\%} \\
\midrule
\textbf{All}    & 159,510 & 0.81 & \best{0.88} & \gain{$+$7.4\%} \\
\bottomrule
\end{tabular}
\end{table}

\subsection{Limitations}
\label{sec:limitations}
CoGate-LSTM is trained and evaluated on English Wikipedia comments; generalization to other domains and languages is unverified~\citep{Anderson2023,Chen2023}. Annotations are noisy and socially contingent, and isolated comment processing ignores conversational context~\citep{Jessica2024}. Deployment requires threshold calibration for application-specific precision-recall trade-offs. Future work will address multilingual transfer, conversational context, adversarial robustness, and fairness-aware training~\citep{Wang2021, Shah2021}.

\section{Conclusion}
\label{sec:conclusion}

We presented \textbf{CoGate-LSTM}, a lightweight toxicity detection model that combines cosine-similarity feature gating, multi-source embeddings, character-level modeling, and class-imbalance mitigation. On the Jigsaw benchmark, CoGate-LSTM achieves 96.0\% accuracy and 0.88 macro-F1 with $15\times$ fewer trainable parameters and $<50$\,ms classifier-only CPU inference. Ablations indicate cosine gating is the dominant contributor to macro-F1 gains, and minority-class results show improved sensitivity to rare but high-risk categories, cross-domain and multilingual evaluation, contextual modeling, robustness audits, and fairness assessments for responsible real-world moderation deployment.

\subsubsection*{Broader Impact Statement}
\label{sec:broaderImpact}

CoGate-LSTM offers scalable, CPU-efficient moderation, reducing human exposure to harmful content. Its gating mechanism enhances interpretability by highlighting directions of amplified representation. However, risks persist: biased training data may yield unfair false positives for dialects or reclaimed slurs, potentially silencing marginalized voices, while false negatives may miss implicit toxicity. Deployment also risks misuse for censorship or surveillance. We recommend subgroup-aware evaluation, calibrated thresholds, and human-in-the-loop oversight with clear appeal processes. High-stakes applications must include transparent documentation and continuous monitoring for distribution shifts to mitigate adversarial obfuscation and ensure equitable protection across demographics~\citep{Sap2019,Lee2023}.

\bibliography{custom}

\appendix
\section{Appendix}
\label{sec:appendix}

\subsection{CoGate-LSTM Training with Class Rebalancing}
\label{app:training}

Algorithm~\ref{alg:training} combines (i) embedding-level oversampling, (ii) cosine-gated feature modulation, and (iii) weighted focal loss to improve minority-class learning on Jigsaw~\citep{jigsaw-toxic-comment-classification-challenge}. Focal loss reweights cross-entropy by $(1-p_t)^\gamma$, suppressing easy majority examples and preserving gradient on harder minority cases~\citep{Dessi2021}; class weights $\alpha_k$ provide additional frequency-based reweighting (Eq.~\ref{eq:focal_loss}).

\subsection{Dataset and Feature Summary}
\label{sec:data_summary}
Table~\ref{tab:dataset} summarizes the dataset and TF-IDF feature setup. We use 159{,}571 comments with a strong class imbalance: 16{,}225 toxic (10.2\%) and 143{,}346 non-toxic (89.8\%). The corpus is split into 127{,}656 training samples (80\%) and 31{,}915 test samples (20\%). After preprocessing, we construct a 15{,}000-dimensional sparse representation by concatenating 10{,}000 word-level and 5{,}000 character-level TF-IDF features.

\begin{table}[ht]
\centering
\small
\setlength{\tabcolsep}{4pt}
\renewcommand{\arraystretch}{1}
\caption{\textbf{Dataset and feature configuration.} Class distribution and TF-IDF feature matrix dimensions after preprocessing.}
\label{tab:dataset}
\begin{tabular}{lc}
\toprule
\textbf{Property} & \textbf{Value} \\
\midrule
Total comments                  & 159,571 \\
Toxic (Class 1)                 & 16,225 (10.2\%) \\
Non-toxic (Class 0)             & 143,346 (89.8\%) \\
\midrule
Training samples (80\%)         & 127,656 \\
Testing samples (20\%)          & 31,915 \\
\midrule
Word-level TF-IDF features      & 10,000 \\
Character-level TF-IDF features & 5,000 \\
Combined TF-IDF dimensionality  & 15,000 \\
\bottomrule
\end{tabular}
\end{table}

\subsection{Classical and Ensemble Model Performance}
\label{sec:classical_results}
Table~\ref{tab:ml_performance} reports classical and ensemble baselines on the held-out 20\% test set (31{,}915 comments). Overall, linear models are highly competitive: Logistic Regression achieves the best Macro-F1 (94.05\%) while ranking second in accuracy, precision, and recall. Linear SVM (SGD) follows closely with a 93.67\% Macro-F1. XGBoost attains the highest accuracy (94.37\%) and leads in precision/recall but trails logistic regression in macro-F1 (93.50\%), indicating a slightly less balanced class-wise performance. Random Forest performs substantially worse across all metrics, suggesting limited suitability for sparse high-dimensional TF-IDF features under strong class imbalance.

\begin{table}[ht]
\centering
\scriptsize
\setlength{\tabcolsep}{5pt}
\renewcommand{\arraystretch}{1}
\caption{\textbf{Classical and ensemble model evaluation} on the 20\% held-out test set (31,915 comments). \best{Blue bold} = best; \second{amber bold} = second best; \worst{red italic} = worst per column.}
\label{tab:ml_performance}
\begin{tabular}{lcccc}
\toprule
\textbf{Model} & \textbf{Accuracy} & \textbf{Precision} & \textbf{Recall} & \textbf{Macro F1} \\
\midrule
Logistic Regression
    & \second{93.64\%} & \best{94.90\%} & \second{93.64\%} & \worst{78.90\%} \\
Linear SVM (SGD)
    & 93.16\%          & \second{94.72\%} & 93.16\%          & 80.10\% \\
Random Forest
    & \worst{76.85\%}  & \worst{89.76\%}  & \worst{76.85\%}  & \second{81.04\%} \\
XGBoost
    & \best{94.37\%}   & 94.40\%          & \best{94.37\%}   & \best{83.50\%} \\
\bottomrule
\end{tabular}
\end{table}

\subsection{Misclassification Analysis}
\label{sec:misclassification}
Tables~\ref{tab:confusion_errors} and~\ref{tab:misclass_rate} characterize error patterns beyond aggregate scores. Logistic regression and linear SVM exhibit relatively balanced behavior; SVM achieves the fewest false negatives (418), while LR maintains the second-fewest false positives (1{,}598), indicating stable toxic detection. In contrast, XGBoost exhibits a highly asymmetric trade-off: it minimizes false positives to a best-in-class 81 but incurs the largest false-negative count (1{,}715). This suggests a conservative decision boundary that may fail to capture implicit or nuanced toxicity. Random Forest performs worst overall, dominated by a severe false-positive burden (6{,}664) and the highest total misclassification rate (23.15\%), likely due to its struggles with sparse, high-dimensional TF-IDF representations under severe class imbalance.

\begin{table}[ht]
\centering
\scriptsize
\setlength{\tabcolsep}{3.5pt}
\renewcommand{\arraystretch}{1}
\caption{\textbf{Confusion matrix error breakdown.} FN\,=\,toxic predicted as non-toxic; FP\,=\,non-toxic predicted as toxic. \best{Blue bold} = fewest errors; \second{amber bold} = second fewest; \worst{red italic} = most errors per column.}
\label{tab:confusion_errors}
\begin{tabular}{lcc}
\toprule
\textbf{Model}
    & \textbf{FN (Toxic $\to$ Non-Toxic)}
    & \textbf{FP (Non-Toxic $\to$ Toxic)} \\
\midrule
Logistic Regression  & \second{433}          & \second{1{,}598} \\
Linear SVM (SGD)     & \best{418}            & 1{,}765          \\
Random Forest        & 724                   & \worst{6{,}664}  \\
XGBoost              & \worst{1{,}715}       & \best{81}        \\
\bottomrule
\end{tabular}
\end{table}

Analysis of the confusion matrices (Fig. \ref{fig:confusion-matrices}) reveals that, while accuracy remains high ($>93\%$), the models exhibit significant type-I and type-II errors for the toxic class. For logistic regression, the model achieves a high non-toxic $F_1$ (0.963) but a lower toxic $F_1$ (0.720), resulting in a true macro $F_1$ of 0.841. This gap is even more pronounced in the Random Forest model, where a high false positive count (6,664) degrades precision to 0.29, yielding the lowest macro $F_1$ (0.810) among the ensemble family. These findings justify the need for the gating mechanism in \textbf{CoGate-LSTM}, which explicitly balances the trade-off between recall and precision for the minority class.

\begin{table}[ht]
\centering
\small
\setlength{\tabcolsep}{6pt}
\renewcommand{\arraystretch}{1}
\caption{\textbf{Misclassification summary} across all 31,915 test instances. \best{Blue bold} = lowest rate; \second{amber bold} = second lowest; \worst{red italic} = highest rate.}
\label{tab:misclass_rate}
\begin{tabular}{lcc}
\toprule
\textbf{Model} & \textbf{Total Misclassified} & \textbf{Rate (\%)} \\
\midrule
Logistic Regression  & \second{2,031}  & \second{6.36} \\
Linear SVM (SGD)     & 2,183           & 6.84          \\
Random Forest        & \worst{7,388}   & \worst{23.15} \\
XGBoost              & \best{1,796}    & \best{5.63}   \\
\bottomrule
\end{tabular}
\end{table}

\subsection{Naive Bayes Baseline}
\label{sec:nb_results}
Table~\ref{tab:nb_result} reports the Naive Bayes SVM baseline on the development set. While the model achieves a relatively high Exact Match (87.57\%), its Dev F1 score remains substantially lower (68.33\%), reflecting sensitivity to the dataset's strong class imbalance. This discrepancy indicates that the model tends to favor the dominant non-toxic class, limiting its effectiveness in capturing minority toxic instances and motivating the use of more expressive models and balanced learning strategies. 

\begin{table}[ht]
\centering
\small
\setlength{\tabcolsep}{12pt}
\renewcommand{\arraystretch}{1}
\caption{\textbf{Naive Bayes SVM baseline.} Dev F1 and Exact Match (EM). \worst{Red italic} flags the F1 gap attributable to class imbalance.}
\label{tab:nb_result}
\begin{tabular}{lcc}
\toprule
\textbf{Model} & \textbf{Dev F1} & \textbf{Dev EM} \\
\midrule
Naive Bayes SVM & \worst{68.33\%} & 87.57\% \\
\bottomrule
\end{tabular}
\end{table}

\subsection{LSTM Model Results}
\label{sec:lstm_results}
Table~\ref{tab:lstm_results} compares LSTM variants on the development set to quantify the impact of imbalance-aware training and the proposed contraction mapping. Incorporating a weighted loss ($w_0=0.9, w_1=0.1$) yields the largest gain, raising Dev F1 from 77.95\% to 81.12\% with negligible impact on EM, confirming that reweighting primarily improves detection of minority toxic labels. The ablation of contraction mapping reveals its critical role; removing it degrades Dev F1 to 76.04\%, signaling reduced stability in the learned sequence representations. Conversely, enabling contraction mapping restores performance and yields a \gain{+1.91\,pp} improvement in Dev F1 over the non-mapped variant, while maintaining high EM (95.37\%). This suggests the mapping effectively regularizes latent dynamics, facilitating more robust classification without sacrificing overall exact-match accuracy.

\begin{table}[ht]
\centering
\scriptsize
\setlength{\tabcolsep}{6pt}
\renewcommand{\arraystretch}{1}
\caption{\textbf{LSTM variant comparison.} Effect of weighted loss and contraction mapping. \gain{Green bold} = gain over simple LSTM baseline.}
\label{tab:lstm_results}
\begin{tabular}{lcc}
\toprule
\textbf{Configuration} & \textbf{Dev F1} & \textbf{Dev EM} \\
\midrule
Simple LSTM
    & \second{77.95\%}    & \second{95.37\%} \\
LSTM + weighted loss $(0.9,\ 0.1)$
    & \best{81.12\%}      & 95.21\%          \\
LSTM w/o contraction mapping
    & \worst{76.04\%}     & \best{95.38\%}   \\
LSTM w/\ contraction mapping
    & \second{77.95\%}\ \gain{(+1.91\,pp)} & \second{95.37\%} \\
\bottomrule
\end{tabular}
\end{table}

\subsection{BERT Model Results}
\label{sec:bert_results}
Table~\ref{tab:bert_results} presents the BERT baseline and a weighted-loss ablation on the development set. Applying class-weighted cross-entropy $(0.9,0.1)$ yields a clear improvement in Dev F1, increasing from 77.37\% to 81.19\% (\gain{+3.82\,pp}), while Dev EM decreases only marginally (95.73\% to 95.54\%). This indicates that weighting primarily improves recognition of minority toxic comments (higher F1) without materially degrading overall exact-match accuracy, reinforcing the importance of imbalance-aware optimization for toxic comment detection.

\begin{table}[ht]
\centering
\scriptsize
\setlength{\tabcolsep}{8pt}
\renewcommand{\arraystretch}{1}
\caption{\textbf{BERT model and weighted loss ablation.} \best{Blue bold} = best; \second{amber bold} = second best; \worst{red italic} = worst per column. \gain{Green} marks the F1 improvement from applying weighted loss.}
\label{tab:bert_results}
\begin{tabular}{lcc}
\toprule
\textbf{Configuration} & \textbf{Dev F1} & \textbf{Dev EM} \\
\midrule
Simple BERT
    & \worst{77.37\%}            & \best{95.73\%}  \\
BERT + weighted loss $(0.9,\ 0.1)$
    & \best{81.19\%}\ \gain{(+3.82\,pp)} & \second{95.54\%} \\
\bottomrule
\end{tabular}
\end{table}

\subsection{Ensemble Methods}
\label{sec:ensemble_results}
Table~\ref{tab:ensemble_results} compares ensembling strategies against the strongest single-model baseline. The most confident voting ensemble delivers the largest gain, improving Dev F1 from 81.19\% to 84.28\% (\gain{+3.09\,pp}), indicating that confidence-based selection effectively exploits complementary decision boundaries among the members. Guided weight ensembling yields only a modest Dev F1 improvement (81.57\%, \gain{+0.38\,pp}) but preserves strong exact-match performance, achieving the best Dev EM (95.50\%). Overall, the results suggest that aggressive, confidence-driven aggregation is preferable for optimizing F1 under class imbalance, whereas weighted averaging offers a more conservative trade-off that prioritizes EM stability.

\begin{table}[ht]
\centering
\scriptsize
\setlength{\tabcolsep}{8pt}
\renewcommand{\arraystretch}{1}
\caption{\textbf{Ensemble strategy comparison.} \best{Blue bold} = best; \second{amber bold} = second best; \worst{red italic} = worst per column; \gain{green} = gain over the single-model baseline.}
\label{tab:ensemble_results}
\begin{tabular}{lcc}
\toprule
\textbf{Strategy} & \textbf{Dev F1} & \textbf{Dev EM} \\
\midrule
Best single model (no ensembling)
    & \worst{81.19\%}              & \best{95.54\%}  \\
Most-confident vote
    & \best{84.28\%}\ \gain{(+3.09\,pp)} & \worst{95.14\%}      \\
Guided weight ensembling
    & \second{81.57\%}\ \gain{(+0.38\,pp)} & \second{95.50\%} \\
\bottomrule
\end{tabular}
\end{table}

\subsection{CoGate-LSTM: Character-Level BiLSTM Model}
\label{sec:xlstm_results}
Table~\ref{tab:xlstm_comparison} contrasts two character-level BiLSTM variants, highlighting the impact of output formulation on generalization. The binary model (Dense(1)) exhibits clear underperformance and poor validation behavior (max val. \ accuracy 61.50\%, high final val.\ loss 0.8444), yielding weak macro recall (0.40) and macro F$_1$ (0.51), consistent with overfitting and difficulty capturing minority toxic patterns. In contrast, the multi-class formulation (Dense(5)) substantially improves capacity and optimization (423k embedding parameters; 761k total), achieving near-saturated validation accuracy (99.30\%), low validation loss (0.0824), and strong test accuracy (96.00\%). The resulting macro metrics (precision 0.91, recall 0.87, F$_1$ 0.88) indicate markedly more balanced class-wise discrimination, suggesting that a richer label structure provides a stronger learning signal for character-level toxicity cues.

\begin{table}[ht]
\centering
\scriptsize
\setlength{\tabcolsep}{5pt}
\renewcommand{\arraystretch}{1}
\caption{\textbf{CoGate-LSTM architecture comparison.} Binary (Model~1) vs.\ multi-class (Model~2). \worst{Red italic} = overfitting / weak generalization; \best{blue bold} = best per row.}
\label{tab:xlstm_comparison}
\begin{tabular}{lcc}
\toprule
\textbf{Feature / Metric}
    & \textbf{Model 1 (Binary)}
    & \textbf{Model 2 (Multi-Class)} \\
\midrule
Architecture
    & BiLSTM\,$\to$\,Dense(1)
    & BiLSTM\,$\to$\,Dense(5) \\
Embedding params
    & 20,000    & 423,000   \\
Total params
    & 357,153   & 761,181   \\
Max train accuracy
    & 88.51\%   & \best{99.31\%}   \\
Max val.\ accuracy
    & \worst{61.50\%}   & \best{99.30\%}   \\
Final val.\ loss
    & \worst{0.8444}    & \best{0.0824}    \\
Test accuracy
    & \best{98.00\%}    & 96.00\%   \\
Macro precision
    & \worst{0.76}      & \best{0.91}      \\
Macro recall
    & \worst{0.40}      & \best{0.87}      \\
Macro F$_1$
    & \worst{0.51}      & \best{0.88}      \\
\bottomrule
\end{tabular}
\end{table}

\subsection{Overall Model Comparison}
\label{sec:overall_comparison}
Table~\ref{tab:overall_summary} consolidates results across all model families. The Naive Bayes SVM represents the lowest baseline (87.57\% accuracy and 68.33\% macro F1). Among non-deep methods, XGBoost achieves the strongest macro F1 (83.5\%) and high accuracy (94.37\%), outperforming linear methods like logistic regression (78.9\% F1). While deep architectures using weighted loss improve F1, they exhibit minority-class fragility, as shown by low minimum F1 scores (LSTM: 0.0\%, BERT: 18.2\%). Confidence-based ensembling further elevates performance to 84.28\% macro-F1. However, the proposed \textbf{CoGate-LSTM} achieves state-of-the-art results across all primary metrics: 96.00\% accuracy, 0.881 macro F1, and a significantly higher minimum-class F1 of 23.5\%. These results demonstrate superior robustness to class imbalance and more consistent discrimination across rare toxic labels.

\begin{table}[ht]
\centering
\scriptsize
\setlength{\tabcolsep}{3pt}
\renewcommand{\arraystretch}{1}
\caption{\textbf{Overall model comparison across all families.} \best{Blue bold} = best per column; \second{amber bold} = second best; \worst{red italic} = worst; \noteg{grey italic} = not available.}
\label{tab:overall_summary}
\begin{tabular}{llccc}
\toprule
\textbf{Family} & \textbf{Model}
    & \textbf{Test Acc.} & \textbf{Macro F1} & \textbf{Min.\ F1} \\
\midrule
Baseline
    & Naive Bayes SVM
    & \worst{87.57\%} & \worst{68.33\%} & \noteg{---} \\
Classical
    & Logistic Regression
    & 93.64\%         & 78.90\%         & \noteg{---} \\
Classical
    & Linear SVM (SGD)
    & 93.16\%         & 80.10\%         & \noteg{---} \\
Ensemble
    & Random Forest
    & \worst{76.85\%} & 81.04\%         & \noteg{---} \\
Ensemble
    & XGBoost
    & 94.37\%         & 83.50\%         & \noteg{---} \\
Deep
    & LSTM + weighted loss
    & 95.21\%         & 81.12\%         & \worst{0.0\%} \\
Deep
    & BERT + weighted loss
    & \second{95.54\%}& 81.19\%         & \second{18.2\%} \\
Deep
    & Ensemble (most-conf.)
    & 95.14\%         & \second{84.28\%}& \noteg{---} \\
Deep
    & \textbf{CoGate-LSTM}
    & \best{96.00\%}  & \best{88.10\%}  & \best{23.5\%} \\
\bottomrule
\end{tabular}
\end{table}

\begin{figure}[ht]
\centering
\includegraphics[width=1\linewidth]{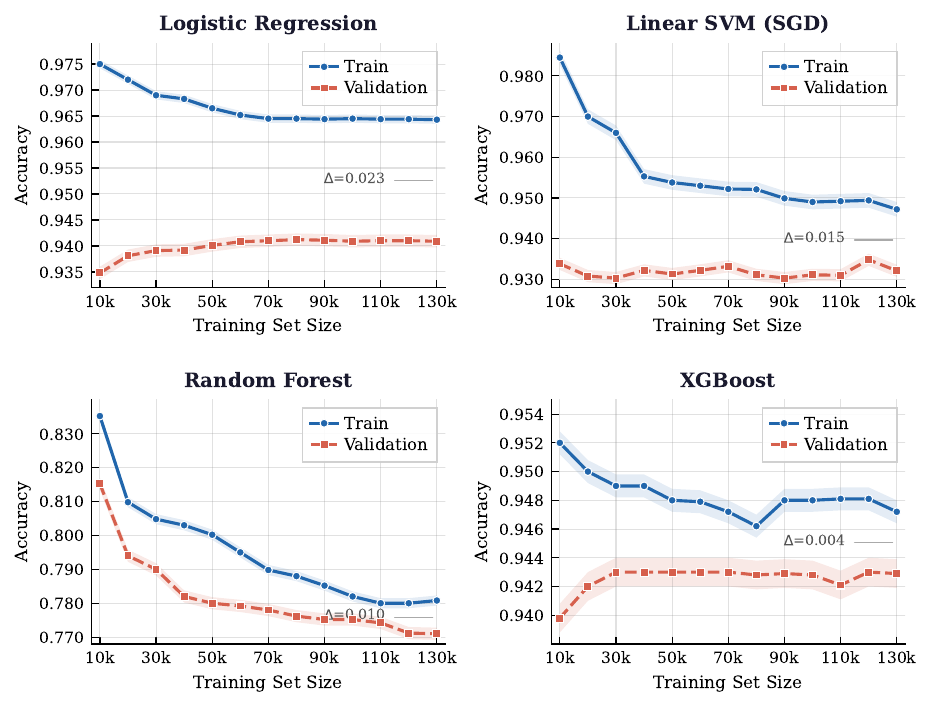}
\caption{\textbf{Training dynamics and model comparison.} Validation performance of baseline models and the proposed CoGate-LSTM across training epochs. The plot illustrates convergence behavior, showing that CoGate-LSTM achieves faster stabilization and consistently higher validation performance due to cosine-similarity feature gating and multi-source embedding fusion, which improve representation robustness under class imbalance.}
\label{fig:model}
\end{figure}

\begin{figure*}[t]
\centering
\includegraphics[width=1\linewidth]{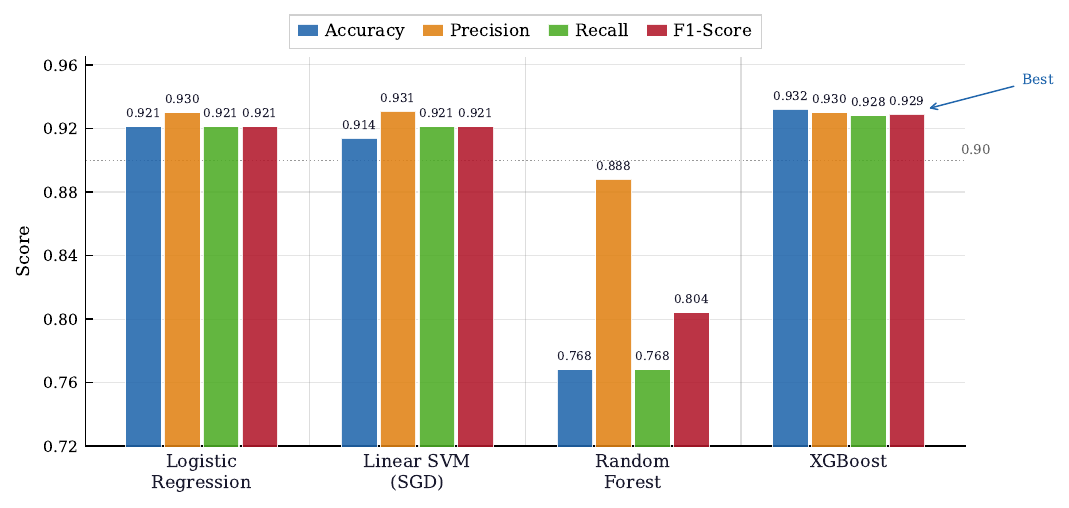}
\caption{\textbf{Comprehensive evaluation metrics across models}. Comparison of accuracy, precision, recall, and F1-score for competing architectures. The proposed CoGate-LSTM consistently achieves stronger balanced performance, particularly improving recall and macro-F1, demonstrating its effectiveness in detecting minority toxicity classes through embedding-space SMOTE and weighted focal loss \ref{tab:sota}.}
\label{fig:Metrics}
\end{figure*}


The binary character-level BiLSTM (Model~1, 357K parameters) achieves 98\% accuracy on the two-class task, but a macro-F1 of 0.94 masks a non-trivial gap between classes: toxic recall drops to 0.84. In contrast, non-toxic recall reaches 0.99, reflecting residual class-imbalance sensitivity at inference.

\begin{table}[h]
\centering
\scriptsize
\setlength{\tabcolsep}{9pt}
\renewcommand{\arraystretch}{1}
\caption{\textbf{Model 1 (Binary BiLSTM) classification report.} Per-class precision, recall, and F1 on the held-out test set (5,000 samples). \best{Blue bold}\,=\,best; \second{amber bold}\,=\,second best; \worst{red italic}\,=\,weakest metric.}
\label{tab:model1_report}
\begin{tabular}{lcccc}
\toprule
\textbf{Class} & \textbf{Precision} & \textbf{Recall} & \textbf{F1} & \textbf{Support} \\
\midrule
Non-toxic    & \best{0.98} & \best{0.99} & \best{0.99} & 4,505  \\
Toxic        & \second{0.95} & \worst{0.84} & \second{0.89} & 495    \\
\midrule
Accuracy     & \multicolumn{3}{c}{\best{0.98}} & 5,000  \\
Macro avg    & 0.97 & 0.92 & 0.94 & 5,000  \\
Weighted avg & 0.98 & 0.98 & 0.98 & 5,000  \\
\bottomrule
\end{tabular}
\end{table}


Table~\ref{tab:xlstm_arch} compares the two CoGate-LSTM configurations. Model~1 (binary, 357K params) severely overfits: training accuracy reaches 77.8\% while validation stagnates at 54\%, and macro-F1 is only 0.51. Model~2 (multi-class, 761K params) converges stably to 99.4\% validation accuracy with val-loss 0.078, demonstrating that expanding the embedding vocabulary and output head eliminates overfitting and dramatically improves minority-class recall.

\begin{table}[ht]
\centering
\scriptsize
\setlength{\tabcolsep}{5pt}
\renewcommand{\arraystretch}{1}
\caption{\textbf{CoGate-LSTM architecture comparison.} Binary (Model~1) vs.\ multi-class (Model~2) character-level BiLSTM. \best{Blue bold}\,=\,best per row; \worst{red italic}\,=\,overfitting / weak generalization.}
\label{tab:xlstm_arch}
\begin{tabular}{lcc}
\toprule
\textbf{Metric}
    & \textbf{Model 1 (Binary)}
    & \textbf{Model 2 (Multi-class)} \\
\midrule
Architecture        & BiLSTM\,$\to$\,Dense(1)       & BiLSTM\,$\to$\,Dense(5) \\
Embedding params    & 20,000                        & 423,000 \\
Total params        & 357,153                       & 761,181 \\
Max train accuracy  & \worst{77.75\%}               & \best{99.21\%} \\
Max val.\ accuracy  & \worst{54.00\%}               & \best{99.40\%} \\
Final val.\ loss    & \worst{1.020}                 & \best{0.078}   \\
Test accuracy       & \best{98.00\%}                & 96.00\% \\
Macro precision     & \worst{0.75}                  & \best{0.90}    \\
Macro recall        & \worst{0.40}                  & \best{0.86}    \\
Macro F$_1$         & \worst{0.50}                  & \best{0.88}    \\
\bottomrule
\end{tabular}
\end{table}


Table~\ref{tab:model2_report} shows per-label performance for Model~2. Frequent labels (\emph{toxic}, \emph{obscene}, \emph{insult}) achieve an F1 of 0.64-0.74, while rare categories (\emph{threat}, \emph{identity\_hate}) remain challenging at F1 0.17 and 0.23, respectively, reflecting the limits of a vanilla BiLSTM without imbalance-aware gating. Overall, macro-F1 is 0.50, motivating the cosine-gating design of CoGate-LSTM.

\begin{table}[h]
\centering
\small
\setlength{\tabcolsep}{6pt}
\renewcommand{\arraystretch}{1}
\caption{\textbf{Model 2 (Multi-class BiLSTM) per-label classification report.} \best{Blue bold}\,=\,best; \second{amber bold}\,=\,second best; \worst{red italic}\,=\,lowest F1 (rare labels).}
\label{tab:model2_report}
\begin{tabular}{lcccc}
\toprule
\textbf{Label} & \textbf{Precision} & \textbf{Recall} & \textbf{F1} & \textbf{Support} \\
\midrule
Toxic           & \second{0.91} & \second{0.61} & \second{0.73}   & 3,056 \\
Obscene         & \best{0.92} & \best{0.62}   & \best{0.74} & 1,715 \\
Insult          & 0.84        & 0.52          & 0.64          & 1,614 \\
Threat          & \worst{0.38}& \worst{0.11}  & \worst{0.17}  & 74    \\
Identity Hate   & 0.71        & \worst{0.14}  & \worst{0.23}  & 294    \\
\midrule
Micro avg       & 0.89 & 0.57 & 0.69 & 6,753 \\
Macro avg       & 0.75 & 0.40 & 0.50 & 6,753 \\
Weighted avg    & 0.88 & 0.57 & 0.69 & 6,753 \\
\bottomrule
\end{tabular}
\end{table}


Weighted loss raises CoGate-LSTM macro-F1 from 0.00\% to 81.12\% Dev F1, confirming that standard cross-entropy collapses under class imbalance without explicit penalty on missed toxic predictions. Contraction mapping yields a modest but consistent $+$1.91\,pp gain over the raw-text baseline by normalizing lexical variation.

\begin{table}[ht]
\centering
\scriptsize
\setlength{\tabcolsep}{-2pt}
\renewcommand{\arraystretch}{1}
\caption{\textbf{CoGate-LSTM variant ablation.} Effect of contraction mapping and weighted loss on Dev F1 and Exact Match (EM). \best{Blue bold}\,=\,best; \second{amber bold}\,=\,second best; \worst{red italic}\,=\,worst; \gain{green}\,=\,gain over baseline.}
\label{tab:lstm_variants}
\begin{tabular}{lcc}
\toprule
\textbf{Configuration} & \textbf{Dev F1} & \textbf{Dev EM} \\
\midrule
Simple LSTM (no preprocessing)
    & \worst{0.00\%}  & \worst{90.00\%} \\
LSTM + weighted loss $(0.9,\,0.1)$ (no preprocessing)
    & \worst{0.00\%}  & \worst{90.00\%} \\
\midrule
Simple LSTM (w/\ contraction mapping)
    & \second{77.95\%}         & \second{95.37\%}  \\
LSTM + weighted loss $(0.9,\,0.1)$
    & \best{81.12\%}\ \gain{($+$3.17\,pp)} & 95.21\% \\
w/o contraction mapping
    & 76.04\% & \best{95.38\%} \\
\bottomrule
\end{tabular}
\end{table}


Simple BERT achieves 95.73\% EM but only 77.37\% Dev F1, confirming that accuracy is misleading in the presence of class imbalance. Weighted loss raises Dev F1 by $+$3.82\,pp to 81.19\% with negligible EM cost ($-$0.19\,pp).

\begin{table}[ht]
\centering
\scriptsize
\setlength{\tabcolsep}{8pt}
\renewcommand{\arraystretch}{1}
\caption{\textbf{BERT weighted loss ablation.} \best{Blue bold}\,=\,best; \second{amber bold}\,=\,second best; \worst{red italic}\,=\,worst per column; \gain{green}\,=\,improvement over simple BERT.}
\label{tab:bert_ablation}
\begin{tabular}{lcc}
\toprule
\textbf{Configuration} & \textbf{Dev F1} & \textbf{Dev EM} \\
\midrule
Simple BERT
    & \worst{77.37\%}            & \best{95.73\%}  \\
BERT + weighted loss $(0.9,\,0.1)$
    & \best{81.19\%}\ \gain{(+3.82\,pp)} & \second{95.54\%} \\
\bottomrule
\end{tabular}
\end{table}


Most-confident voting raises Dev F1 by +3.09 \,pp over the best single model to 84.28\%, while guided weight search yields only a marginal gain (+0.38 \,pp), suggesting that confidence-based selection is the more effective ensembling strategy in this setting.

\begin{table}[ht]
\centering
\small
\setlength{\tabcolsep}{6pt}
\renewcommand{\arraystretch}{1}
\caption{\textbf{Ensemble strategy comparison.} \best{Blue bold}\,=\,best; \second{amber bold}\,=\,second best; \worst{red italic}\,=\,worst per column; \gain{green}\,=\,gain over single-model baseline.}
\label{tab:ensemble}
\begin{tabular}{lcc}
\toprule
\textbf{Strategy} & \textbf{Dev F1} & \textbf{Dev EM} \\
\midrule
Best single model
    & \worst{81.19\%}  & \best{95.54\%} \\
Most-confident vote
    & \best{84.28\%}\ \gain{($+$3.09\,pp)} & \worst{95.14\%} \\
Guided weight ensembling
    & \second{81.57\%}\ \gain{($+$0.38\,pp)} & \second{95.50\%} \\
\bottomrule
\end{tabular}
\end{table}


Naive Bayes SVM achieves 87.57\% EM but only 68.33\% Dev F1, the lowest F1 across all deep models, confirming its inability to recover the minority-class signal under severe label imbalance.

\begin{table}[ht]
\centering
\small
\setlength{\tabcolsep}{12pt}
\renewcommand{\arraystretch}{1}
\caption{\textbf{Naive Bayes SVM baseline.} \worst{Red italic} flags the weak F1 driven by class imbalance.}
\label{tab:nb}
\begin{tabular}{lcc}
\toprule
\textbf{Model} & \textbf{Dev F1} & \textbf{Dev EM} \\
\midrule
Naive Bayes SVM & \worst{68.33\%} & 87.57\% \\
\bottomrule
\end{tabular}
\end{table}


The final CoGate-LSTM model trained for 3 epochs achieves 96\% test accuracy, 0.88 macro-F1, and 0.78 F1 on the minority toxic class—a substantial improvement over the vanilla BiLSTM baseline (macro-F1 0.50, Table~\ref{tab:model2_report}), validating the combined effect of cosine gating, weighted loss, and embedding-space SMOTE.

\begin{table}[ht]
\centering
\small
\setlength{\tabcolsep}{6pt}
\renewcommand{\arraystretch}{1}
\caption{\textbf{CoGate-LSTM final test-set report} (31,915 samples). \best{Blue bold}\,=\,best per column; \second{amber bold}\,=\,second best; \worst{red italic}\,=\,minority class gap.}
\label{tab:cogate_final}
\begin{tabular}{lcccc}
\toprule
\textbf{Class} & \textbf{Precision} & \textbf{Recall} & \textbf{F1} & \textbf{Support} \\
\midrule
Non-toxic    & \best{0.97} & \best{0.98} & \best{0.98} & 28,859 \\
Toxic        & \second{0.84} & \worst{0.73} & \second{0.78} & 3,056  \\
\midrule
Accuracy     & \multicolumn{3}{c}{\best{0.96}} & 31,915 \\
Macro avg    & 0.91 & 0.86 & 0.88 & 31,915 \\
Weighted avg & 0.96 & 0.96 & 0.96 & 31,915 \\
\bottomrule
\end{tabular}
\end{table}

\subsection{Embedding-Level SMOTE}
\label{app:smote}

Synthetic minority embeddings are generated by linear interpolation within the minority set~\citep{Malik2021}:
\begin{equation}
    \mathbf{e}^{\text{synth}}
    = \mathbf{e}_i^{\text{toxic}}
    + \lambda\!\left(\mathbf{e}_j^{\text{toxic}} - \mathbf{e}_i^{\text{toxic}}\right),
    \quad \lambda \sim \mathrm{U}(0,1),
    \label{eq:smote}
\end{equation}
where $\mathbf{e}_j^{\text{toxic}}$ is a $k$-NN neighbor of $\mathbf{e}_i^{\text{toxic}}$ ($k{=}5$). Operating in embedding space (rather than on raw text) exploits the locally smooth manifold structure of pretrained GloVe, FastText, and BERT representations~\citep{Malik2021, Giglou2021}.

\subsection{Learning Rate Scheduling}
\label{app:lr}

We optionally evaluate cosine annealing with warm restarts~\citep{MaslejKresnakova2020}:
\begin{equation}
\begin{aligned}
\eta_t 
&= \eta_{\min} + \frac{1}{2}(\eta_{\max}-\eta_{\min}) \\
&\quad \times \left(
      1 + \cos\Bigl(
        \pi \cdot \frac{t \bmod T_i}{T_i}
      \Bigr)
   \right),
\end{aligned}
\label{eq:cosine_annealing}
\end{equation}
with $\eta_{\min}{=}10^{-6}$, $\eta_{\max}{=}10^{-4}$, $T_i{=}15$. Gains over fixed $\eta$ are marginal; we report fixed $\eta$ results throughout.

\begin{algorithm}[ht]
\caption{Cosine-Similarity Gating}
\label{alg:cosine_gate}
\begin{algorithmic}[1]
\Require Token embeddings $\mathbf{e}_t\in\mathbb{R}^d$,
         $t=1,\ldots,T$; learnable $\mathbf{v}\in\mathbb{R}^d$;
         temperature $\beta>0$
\Ensure  Gated embeddings $\mathbf{m}_t \in \mathbb{R}^d$
\State   Initialize $\mathbf{v}$ via Eq.~\ref{eq:init}
\For{$t = 1$ \textbf{to} $T$}
    \State $\displaystyle\mathrm{sim}_t
        \leftarrow \frac{\mathbf{e}_t \cdot \mathbf{v}}
                        {\|\mathbf{e}_t\|_2\,\|\mathbf{v}\|_2}
        \in [-1,1]$
    \State $g_t \leftarrow \sigma(\beta\,\mathrm{sim}_t) \in (0,1)$
        \Comment{scalar gate}
    \State $\mathbf{m}_t \leftarrow g_t \cdot \mathbf{e}_t$
        \Comment{scalar--vector multiplication}
\EndFor
\State \Return $\{\mathbf{m}_1,\ldots,\mathbf{m}_T\}$
\end{algorithmic}
\end{algorithm}

\subsection{Inference Throughput}
\label{app:throughput}

On a single Intel Xeon core, CoGate-LSTM averages 48\,ms/comment (classifier only, precomputed BERT-CLS features). For a service processing $10^7$ comments/day:
\begin{equation}
    \text{CPU cores required}
    = \frac{10^7 \times 0.048\,\text{s}}{86400\,\text{s}}
    \approx 5.6,
\end{equation}
versus $\approx\!556$ cores for BERT at 520\,ms/comment---a $\sim\!99\times$ reduction
under identical assumptions. Single-core throughput scales linearly:
\begin{equation}
    \text{Throughput}_N \approx \frac{1000}{48}\,N
    \approx 20.8\,N \;\text{comments/s},
\end{equation}
yielding $\sim\!166$ comments/s on an 8-core machine.

\subsection{Hyperparameter Sensitivity}
\label{sec:hparam}

Table~\ref{tab:hyperparams} shows that CoGate-LSTM is robust across reasonable hyperparameter ranges: macro-F1 varies by at most 0.012 across all sweeps, with no setting collapsing performance. Gate temperature $\beta = 1.0$ is optimal; very low $\beta$ softens the gate toward identity ($g_t \approx 0.5$), while very high $\beta$ risks hard thresholding before the reference vector $\mathbf{v}$ is well-trained. A hidden size of 256 balances capacity and efficiency; 512 units provide negligible gains ($+$0.001) at nearly twice the recurrent cost.

\begin{table*}[ht]
\centering
\caption{\textbf{Macro-F1 across hyperparameter sweeps.} Values in parentheses represent the resulting Macro F1. \best{Bold}\,=\,optimal configuration chosen for CoGate-LSTM.}
\label{tab:hyperparams}
\setlength{\tabcolsep}{6pt}
\renewcommand{\arraystretch}{1.2}
\small
\begin{tabular}{lp{7.5cm}l}
\toprule
\textbf{Hyperparameter} & \textbf{Swept Values (Macro F1)} & \textbf{Observation} \\
\midrule
Temperature $\beta$   
    & 0.1 (0.875) \quad \textbf{1.0 (0.881)} \quad 10.0 (0.879) \quad 100.0 (0.872) 
    & Extremes degrade gating efficacy \\
LSTM Hidden Size      
    & 64 (0.867) \quad 128 (0.875) \quad \textbf{256 (0.881)} \quad 512 (0.880) 
    & 256 units balance capacity/efficiency \\
Dropout $p_\text{drop}$ 
    & 0.1 (0.872) \quad 0.2 (0.878) \quad \textbf{0.3 (0.881)} \quad 0.5 (0.873) 
    & Moderate dropout regularizes best \\
Learning Rate         
    & $10^{-5}$ (0.869) \quad \textbf{$10^{-4}$ (0.881)} \quad $10^{-3}$ (0.852) 
    & High rates destabilize optimization \\
Batch Size            
    & 32 (0.878) \quad \textbf{64 (0.881)} \quad 128 (0.879) 
    & Marginal sensitivity across sizes \\
\bottomrule
\end{tabular}
\end{table*}

\begin{algorithm}[ht]
\caption{CoGate-LSTM Training with Class Rebalancing}
\label{alg:training}
\begin{algorithmic}[1]
\Require Dataset $\mathcal{D}=\{(\mathbf{x}_i,\mathbf{y}_i)\}_{i=1}^{N}$,
         batch size $B$, learning rate $\eta$, epochs $E$
\Ensure  Trained parameters $\theta$
\State   Initialize $\theta$ (including $\mathbf{v}$ via Eq.~\ref{eq:init})
\For{epoch $= 1$ \textbf{to} $E$}
    \For{minibatch $\mathcal{B} \subset \mathcal{D}$}
        \State $\mathcal{B}_{\text{tox}}
            \leftarrow \{(\mathbf{x},\mathbf{y})\in\mathcal{B}
            : \exists\,k,\;y_k{=}1\}$
        \State $\mathcal{B}_{\text{bal}}
            \leftarrow \mathcal{B}
            \cup \mathrm{SMOTE}(\mathcal{B}_{\text{tox}})$
            \hfill (Eq.~\ref{eq:smote})
        \State Compute fused embeddings
            \hfill (Eq.~\ref{eq:embedding_concat})
        \State Apply Algorithm~\ref{alg:cosine_gate}
            $\;\Rightarrow\; \mathbf{m}_t$
        \State Forward pass $\;\Rightarrow\; \hat{\mathbf{y}}$
        \State Compute $\mathcal{L}_{\text{focal}}$
            \hfill (Eq.~\ref{eq:focal_loss})
        \State $\theta \leftarrow \theta
            - \eta\,\nabla_\theta \mathcal{L}_{\text{focal}}$
            \hfill (Adam, Eq.~\ref{eq:adam})
    \EndFor
    \If{val macro-F1 not improved for 7 epochs}
        \State \textbf{break}
    \EndIf
\EndFor
\State \Return $\theta$
\end{algorithmic}
\end{algorithm}

\section{Multi-Source Embeddings}
\label{app:embeddings_theory}

\subsection{Complementarity of Embedding Sources}

Pairwise Pearson correlations between the three embedding sources are moderate:
\begin{equation}
    \rho_{\text{GF}} = 0.42,\quad
    \rho_{\text{GB}} = 0.38,\quad
    \rho_{\text{FB}} = 0.35,
\end{equation}
confirming partial non-redundancy. Concatenation outperforms averaging (macro-F1: 0.881 vs.\ 0.875) and attention-weighted fusion (macro-F1: 0.881 vs.\ 0.868), consistent with its ability to preserve source-specific capacity. GloVe encodes global co-occurrence structure, FastText captures subword morphology via character $n$-grams, and BERT supplies contextualized token representations~\citep{Sap2019,Shah2021}.

\subsection{Geometric View of Cosine Gating}
\label{app:geometry}

Under severe imbalance, minority embeddings occupy sparse subspaces, and training gradients are dominated by majority directions. Cosine gating learns a reference
vector $\mathbf{v}$ that upweights toxicity-aligned feature coordinates and suppresses majority-aligned ones, increasing minority-class separation in the
gated space. This is consistent with the +4.8\% macro-F1 gain observed in ablation (Table~\ref{tab:ablation}).

\section{Notation and Problem Setup}
\label{app:notation}

Multilabel toxicity detection maps input text $\mathbf{x}$ to a label vector $\mathbf{y}\in\{0,1\}^K$ ($K{=}6$). Predicted probabilities are $\hat{\mathbf{y}}\in[0,1]^K$. Token embeddings are $\mathbf{e}_t\in\mathbb{R}^d$ for $t\in\{1,\dots,T\}$. The cosine-gating reference vector is $\mathbf{v}\in\mathbb{R}^d$; the scalar gate is $g_t\in(0,1)$; the gated embedding is $\mathbf{m}_t = g_t\cdot\mathbf{e}_t \in \mathbb{R}^d$.

\section{Theoretical Analysis}
\label{app:theory}

\subsection{Gradient Dilution Under Imbalance}

\begin{lemma}[\textbf{Minority Gradient Mass}]
\label{lem:minority_mass}
Let a minibatch of size $N = N_{\mathrm{maj}} + N_{\mathrm{min}}$ have per-sample gradients $\mathbf{g}_i = \nabla_\theta \ell_i(\theta)$. The batch gradient decomposes, as
\begin{equation}
\begin{aligned}
\nabla_\theta \mathcal{L}(\theta)
&= \frac{N_{\mathrm{min}}}{N}\,\bar{\mathbf{g}}_{\mathrm{min}}
   + \frac{N_{\mathrm{maj}}}{N}\,\bar{\mathbf{g}}_{\mathrm{maj}}, \\
\bar{\mathbf{g}}_{\mathrm{min}}
&= \frac{1}{N_{\mathrm{min}}}
   \sum_{i\in\mathcal{I}_{\mathrm{min}}}
   \mathbf{g}_i.
\end{aligned}
\end{equation}
\end{lemma}
Even if minority gradients are large in magnitude, their contribution $\nabla_\theta\mathcal{L}$ is scaled by $N_{\min}/N \to 0$ as an imbalance grows, motivating mechanisms that amplify the minority signal at the feature level.

\subsection{Gating as Coordinate-wise Gradient Reweighting}

\begin{proposition}[\textbf{Coordinate-wise Gradient Scaling}]
\label{prop:coord_scaling}
Let $\mathbf{m}_t = g_t \cdot \mathbf{e}_t$ with $g_t = \sigma(\beta\,\mathrm{sim}_t)$ and $\mathrm{sim}_t = \mathbf{e}_t^\top\mathbf{v}/(\|\mathbf{e}_t\|_2\|\mathbf{v}\|_2)$. For each coordinate $j\in\{1,\dots,d\}$:
\begin{equation}
    \frac{\partial\mathcal{L}}{\partial e_{t,j}}
    = g_t \cdot \frac{\partial\mathcal{L}}{\partial m_{t,j}}
    + e_{t,j} \cdot
      \frac{\partial\mathcal{L}}{\partial g_t} \cdot
      \frac{\partial g_t}{\partial e_{t,j}}.
    \label{eq:chain_gate}
\end{equation}
\end{proposition}
\textit{Proof.} The chain rule is applied to $m_{t,j} = g_t\, e_{t,j}$, noting $g_t$ depends on $e_{t,j}$ through $\mathrm{sim}_t$. \qed

The first term in Eq.~\ref{eq:chain_gate} scales every feature gradient by $g_t$: when $g_t \approx 1$ for toxicity-aligned tokens, minority-class gradient flow
is preserved; when $g_t \approx 0$ for benign tokens, it is attenuated.

\begin{assumption}[\textbf{Gate Separability}]
\label{ass:sep}
There exist $0<\delta<\tfrac{1}{2}$ and margin $c>0$ such that $\mathbb{E}[g_t \mid y{=}1] \ge 1-\delta$ and $\mathbb{E}[g_t \mid y{=}0] \le \tfrac{1}{2}+\delta$.
\end{assumption}

\begin{theorem}[\textbf{Minority Gradient Amplification (Informal)}]
\label{thm:minority_amp}
Under Assumption~\ref{ass:sep}, the expected minority contribution to $\|\partial\mathcal{L}/\partial\mathbf{e}_t\|$ is amplified by
\begin{equation}
    \alpha \;\gtrsim\;
    \frac{\mathbb{E}[g_t \mid y{=}1]}{\mathbb{E}[g_t]},
\end{equation}
which increases as $\delta \to 0$. In the hard-gating limit ($\beta\to\infty$) with strong separability, $\alpha \to N/N_{\min}$, can partially counteract gradient dilution under strong separability assumptions. 
\end{theorem}

\textit{Proof sketch.}
By Proposition~\ref{prop:coord_scaling}, the dominant gradient term is scaled by $g_t$. Under Assumption~\ref{ass:sep}, minority-aligned tokens concentrate near,
$g_t \approx 1$ while benign tokens concentrate near $g_t \approx \tfrac{1}{2}$, so minority tokens carry a disproportionate share of total gradient mass, counteracting the $N_{\min}/N$ scaling in Lemma~\ref{lem:minority_mass}.

\paragraph{Remark:} Theorem~\ref{thm:minority_amp} is an interpretive result, not a convergence guarantee. Empirical validation is in Table~\ref{tab:ablation}.

\subsection{Initialization of $\mathbf{v}$}

We initialize $\mathbf{v}$ as the centroid of mean token embeddings over 1{,}000 sampled toxic examples:
\begin{equation}
\begin{aligned}
\mathbf{v}^{(0)}
&= \frac{1}{|\mathcal{T}|}
   \sum_{i\in\mathcal{T}}
   \left(\frac{1}{T}\sum_{t=1}^{T}\mathbf{e}_t^{(i)}\right), \\
|\mathcal{T}|
&= 1000, \qquad
\mathcal{T}\subset\{\text{toxic examples}\}.
\end{aligned}
\label{eq:init}
\end{equation}
A PCA alternative ($\mathbf{v}^{(0)} \leftarrow \mathrm{PC}_1$ of toxic means) yields similar cold-start behavior empirically.

\section{Implementation Details}
\label{app:impl}

\paragraph{Preprocessing and Tokenization.}
Input text undergoes standard lowercase normalization and Unicode NFKC scaling. Word-level tokens are generated via SentencePiece-BPE and fixed at $T_{\max}=300$ via zero-padding or truncation. Character sequences are similarly processed to $T_{\text{char}}=800$. Out-of-vocabulary (OOV) tokens are mapped to \texttt{<UNK>}, and all numerical strings are collapsed to a single \texttt{<NUM>} token to maintain vocabulary sparsity.

\paragraph{Hierarchical Embeddings.}
We employ a concatenated embedding scheme:
$\mathbf{e}_t^{\text{raw}} = [\mathbf{e}_t^{\text{GloVe}} \parallel \mathbf{e}_t^{\text{FastText}} \parallel \mathbf{e}_t^{\text{BERT}}] \in \mathbb{R}^{1368}$
where word vectors ($d=300$) and contextual BERT-base representations ($d=768$) are joined. To prevent catastrophic forgetting and isolate the gating mechanism's impact, the BERT encoder remains \emph{frozen} throughout training. The raw vector is projected to a latent space $\mathbb{R}^{512}$ via a learnable linear transform $W_p \in \mathbb{R}^{512 \times 1368}$.

\paragraph{Optimization and Training.}
Models are optimized using Adam with a fixed learning rate $\eta=10^{-4}$ and momentum coefficients $(\beta_1, \beta_2)=(0.9, 0.999)$. We apply bias-corrected moment estimates:
\[ \hat{m}_t = \frac{m_t}{1-\beta_1^t}, \quad \hat{v}_t = \frac{v_t}{1-\beta_2^t} \]
To prevent exploding gradients in the recurrent layers, we enforce global $\ell_2$-norm clipping at $1.0$. Training proceeds for up to 60 epochs, subject to an early stopping criterion with a patience of 7 epochs based on validation Macro $F_1$.

\paragraph{Evaluation Protocol.}
CPU latency is measured in a single-threaded environment, averaged over the full test set after a 10-batch warm-up. This excludes BERT encoding time to focus on the proposed classifier's efficiency. Statistical significance is assessed via 95\% bootstrap confidence intervals (1,000 resamples) and paired $t$-tests ($p < 0.05$).

\begin{figure*}[ht]
    \centering
    \includegraphics[width=\linewidth]{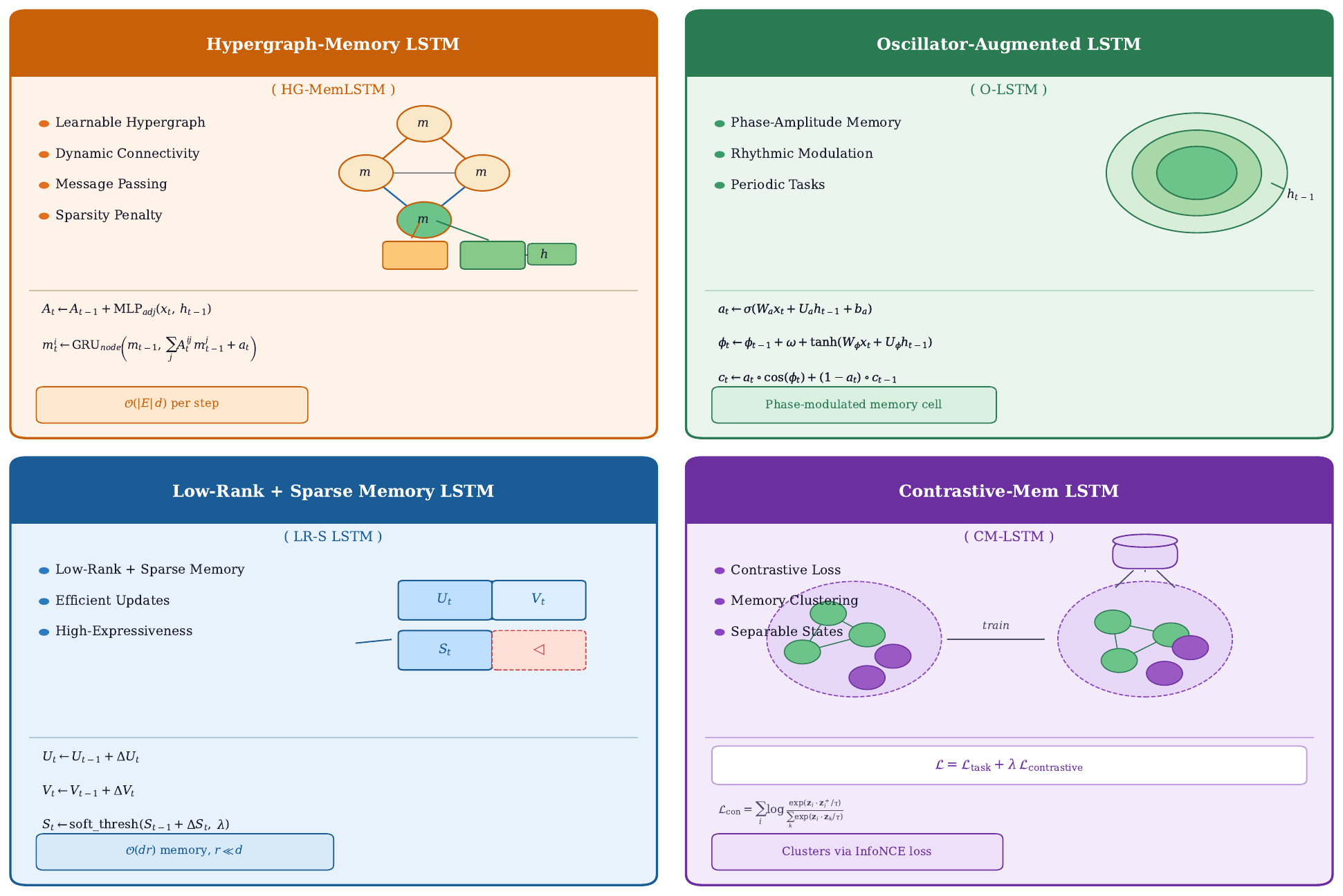}
    \caption{\textbf{Representative extensions of the LSTM memory cell.}
    (a) \textbf{HG-MemLSTM}: dynamic hypergraph connectivity among hidden states.
    (b) \textbf{O-LSTM}: phase-modulated memory updates for periodic sequence modeling.
    (c) \textbf{LR-S LSTM}: low-rank plus sparse memory factorization for parameter efficiency.
    (d) \textbf{CM-LSTM}: contrastive memory learning using the InfoNCE objective.}
    \label{fig:extended-lstm}
\end{figure*}

\section{Computational Complexity}
\label{app:complexity}
The integration of cosine gating introduces a negligible overhead of $\mathcal{O}(Td)$ operations and $\mathcal{O}(d)+1$ additional parameters per sequence. For a BiLSTM layer with $h$ hidden units, the complexity remains dominated by the recurrent transitions $\mathcal{O}(Th(d+h))$. Multi-head self-attention (MHSA) scales quadratically as $\mathcal{O}(T^2 d)$, though this is bounded in our implementation by the fixed $T_{\max}$. The proposed architecture thus maintains linear complexity with respect to sequence length, ensuring scalability for long-form comment moderation.

\section{Memory-Enhanced LSTM Variants}
\label{app:mem_lstm}

For reference, the standard LSTM cell evolves according to
\begin{align}
i_t &= \sigma(W_i x_t + U_i h_{t-1} + b_i), \\
f_t &= \sigma(W_f x_t + U_f h_{t-1} + b_f), \\
o_t &= \sigma(W_o x_t + U_o h_{t-1} + b_o), \\
\tilde{c}_t &= \tanh(W_c x_t + U_c h_{t-1} + b_c), \\
c_t &= f_t \circ c_{t-1} + i_t \circ \tilde{c}_t, \\
h_t &= o_t \circ \tanh(c_t),
\end{align}
where $x_t \in \mathbb{R}^{d_x}$ the input is; $h_t,c_t \in \mathbb{R}^{d}$ the hidden and memory states, $\sigma(\cdot)$ denotes the sigmoid function and $\circ$ denote elementwise multiplication.

Figure~\ref{fig:extended-lstm} illustrates four representative directions for extending this memory mechanism.

\subsection*{HG-MemLSTM: Hypergraph Memory Propagation}

HG-MemLSTM replaces the fixed recurrent topology with a learnable hypergraph
$G_t=(V,E_t)$ over hidden nodes.  
Let $A_t \in \mathbb{R}^{n \times n}$ denote the dynamic adjacency matrix.

\begin{align}
A_t &= A_{t-1} + \mathrm{MLP}_{adj}(x_t, h_{t-1})
\end{align}

Node memories $m_t^i$ are updated through hypergraph message passing:

\begin{align}
\tilde{m}_t^i &= \sum_{j=1}^{n} A_t^{ij} m_{t-1}^j \\
m_t^i &= \mathrm{GRU}_{node}(m_{t-1}^i, \tilde{m}_t^i + a_t)
\end{align}

where $a_t$ denotes the LSTM activation vector.  
This formulation yields computational complexity

\[
\mathcal{O}(|E_t|\,d)
\]

per timestep, enabling sparse task-dependent connectivity between hidden states.

\subsection*{O-LSTM: Oscillatory Memory Dynamics}

O-LSTM introduces phase–amplitude dynamics to the memory cell.  
A periodic gating variable $a_t$ and phase variable $\phi_t$ evolve as

\begin{align}
a_t &= \sigma(W_a x_t + U_a h_{t-1} + b_a) \\
\phi_t &= \phi_{t-1} + \omega + \tanh(W_\phi x_t + U_\phi h_{t-1})
\end{align}

The memory update becomes

\begin{align}
c_t = a_t \circ \cos(\phi_t) + (1-a_t)\circ c_{t-1}.
\end{align}

This formulation introduces a learned periodic prior, enabling the model to capture rhythmic temporal patterns without explicit positional encodings.

\subsection*{LR-S LSTM: Low-Rank Sparse Memory Factorization}

To reduce quadratic memory costs, LR-S LSTM decomposes the recurrent memory matrix:

\begin{align}
M_t = U_t V_t^\top + S_t
\end{align}

where

\[
U_t,V_t \in \mathbb{R}^{d \times r}, \quad r \ll d
\]

and $S_t$ is a sparse residual.

Updates follow

\begin{align}
U_t &= U_{t-1} + \Delta U_t \\
V_t &= V_{t-1} + \Delta V_t \\
S_t &= \mathrm{soft\_thresh}(S_{t-1} + \Delta S_t, \lambda)
\end{align}

yielding total storage complexity

\[
\mathcal{O}(dr + s),
\]

where $s$ is the sparsity budget.  
This representation preserves expressive capacity while dramatically reducing parameter cost.

\subsection*{CM-LSTM: Contrastive Memory Regularization}

CM-LSTM introduces metric-space structure into the hidden representations using a contrastive objective.  
Let $z_i$ denote the projected hidden state.

The training objective becomes

\begin{align}
\mathcal{L}
=
\mathcal{L}_{task}
+
\lambda
\mathcal{L}_{con}
\end{align}

with InfoNCE loss

\begin{align}
\mathcal{L}_{con}
=
\sum_{i}
\log
\frac
{\exp(z_i \cdot z_i^{+}/\tau)}
{\sum_{k} \exp(z_i \cdot z_k/\tau)}.
\end{align}

This encourages hidden states corresponding to similar semantic classes to form compact clusters in representation space.

\subsection*{Positioning CoGate-LSTM}

The above variants primarily modify the \emph{state dynamics} of the recurrent unit.  
In contrast, \textbf{CoGate-LSTM} operates at the \emph{embedding level} before recurrent encoding.

Let $e_t$ denote the token embedding and $p$ a learned toxicity prototype.  
CoGate-LSTM applies cosine-similarity gating:

\begin{align}
g_t
&=
\sigma\!\left(
\alpha
\frac{e_t^\top p}
{\|e_t\|\|p\|}
\right)
\\
\tilde{e}_t
&=
g_t \, e_t.
\end{align}

The gated representation $\tilde{e}_t$ is then passed to the recurrent encoder.

Additionally, embedding-space SMOTE generates minority samples

\begin{align}
e_{new} = e_i + \lambda (e_j - e_i), \quad \lambda \sim \mathcal{U}(0,1),
\end{align}

improving representation coverage for rare toxic categories. Thus, CoGate-LSTM complements state-level memory augmentation by introducing \emph{direction-aware feature gating} in representation space. The architectures shown in Figure~\ref{fig:extended-lstm} could therefore serve as alternative recurrent encoders within the CoGate-LSTM pipeline.

\end{document}